\documentclass[conference]{IEEEtran}
\IEEEoverridecommandlockouts

\usepackage{mathtools}
\usepackage{lineno}
\modulolinenumbers[5]
\usepackage{comment}
\usepackage{graphicx}
\usepackage{amsfonts}
\usepackage{amsmath}
\usepackage{amssymb}
\usepackage{commath}
\usepackage{siunitx}

\usepackage{placeins}
\usepackage{multirow}
\usepackage{array}
\usepackage[linesnumbered,lined,ruled]{algorithm2e}
\usepackage[misc,geometry]{ifsym}
\usepackage{mathtools,lipsum}
\usepackage{amsmath,amssymb,amsfonts}
\usepackage{algorithmic}
\usepackage{graphicx}
\usepackage{textcomp}
\usepackage{xcolor}
\def\BibTeX{{\rm B\kern-.05em{\sc i\kern-.025em b}\kern-.08em
    T\kern-.1667em\lower.7ex\hbox{E}\kern-.125emX}}
\DeclareMathOperator{\asin}{arcsin}

\DeclareMathOperator{\atantwo}{atan2}
\usepackage[font=footnotesize]{subcaption}
\usepackage[font=footnotesize]{caption}
\usepackage{cuted}
\usepackage[capitalise]{cleveref}

\begin{document}

\title{Feedback Motion Plan Verification for Vehicles with Bounded Curvature Constraints\\
\thanks{This research was supported under NASA cooperative agreement number NND13AB04A.
}
}

\author{\IEEEauthorblockN{Giovanni Miraglia, Loyd R. Hook}
\IEEEauthorblockA{\textit{Electrical and Computer Engineering} \\
\textit{The University of Tulsa}\\
Tulsa, Oklahoma, USA \\
giovmi@utulsa.edu; loyd-hook@utulsa.edu}
}

\IEEEoverridecommandlockouts
\IEEEpubid{\makebox[\columnwidth]{978-1-7281-3885-5/19/\$31.00~\copyright2019 IEEE \hfill} \hspace{\columnsep}\makebox[\columnwidth]{ }}

\maketitle
\IEEEpubidadjcol

\begin{abstract}
The kinematic approximation of Dubin’s Vehicle has been largely exploited in the formulation of various motion planning methods. In the majority of these methods, planning and control phases are decoupled, and the burden of rejecting disturbances is left to the controller. An alternative to this approach is the use of a feedback motion plan, where for each state there is a specific pre-computed action that will be executed. This planning approach provides the ability to verify all trajectories off-line. The verification can be performed using backward reachability, which provides the set of configurations from which a region is reachable. In this paper, we formulate a verification process that relies on the computation of the backward reachable set using geometric principles. In addition to the theoretical foundation of the method, we provide a numerical implementation of the method and we illustrate a practical example.
\end{abstract}

\begin{IEEEkeywords}
Autonomous vehicles; motion analysis; motion planning; verification; reachability analysis.
\end{IEEEkeywords}

\section{Introduction}
Currently, we are witnessing rapid growth in the use of autonomy in both ground and aerial vehicles. Many high-level challenges relating to the two types of vehicles can be solved with similar solutions. For instance, one of the most crucial tasks, motion planning, can be simplified using the well-known kinematic approximation of Dubin's Vehicle \cite{Dubins1957} for both ground and aerial vehicles operating at a constant altitude. Various motion planning methods relying on this approximation have been formulated \cite{Lin2014,Chitsaz2007,Macharet2011,Shanmugavel2010}. In the majority of these methods, planning and control phases are decoupled, and the burden of rejecting disturbances is left to the controller. However, unpredictable events may lead to excessive deviations that should not be managed by the controller alone. In these scenarios, an on-line replan is usually required to find a new appropriate path. In many instances this is perfectly acceptable; however, for applications such as those involving unmanned or autonomous air vehicle systems, planned paths are required to be pre-verified in order to assure compliance with regulations and safety constraints and thus cannot be changed on-line. One solution to this conflict is the use of a feedback motion plan \cite{Konkimalla1999,Miraglia2017,Mira1910:Feedback,Goncalves2010a}, where for each state there is a specific pre-computed action that will be executed. This planning approach provides the ability to reject disturbances, recover from deviations, and the possibility to verify all trajectories off-line. 

A feedback motion plan can be continuous or discrete. While for continuous plans, it is possible to prove properties such as stability using classical theorems, for the discrete case the problem is much harder. For this reason, a common practice to perform verification of discrete plans is to use Monte Carlo simulations \cite{Margellos2013}. However, the drawback is that this is a probabilistic approach in which a specific trajectory is tested with a certain probability. Instead in this paper, we formulate a verification method based on geometric principles that allow for an exhaustive analysis of the whole 3-D configuration space. This allows several advantages including verification and subsequent approval of a large number of paths simultaneously, the ability to compare different plans, and the possibility to optimize a plan over specific metrics.

The method that we developed to perform this analysis is based on the use of reachability\cite{Althoff2015,Asarin2006, Mitchell2007, Nilsson2014, Shubert2018}, which is employed in the verification of hybrid and switched systems. There exist two types of reachability: forward \cite{El-Guindy2017} and backward \cite{Xue2016, Adzkiya2014,Pek2018}. Our method combines both backward and forward reachable sets that are obtained exploiting geometric properties. In particular, a time-independent backward propagation of 2-D sets is performed to study the backward reachability of a region of interest.

The paper illustrates the geometric principles underlying the proposed method along with practical implementation. In particular, Section II provides a formal definition of the problem. In Section III, we illustrate how to derive the \emph{Cellular Backward Reachability}, which provides the set of configurations from which a specific border of a cell is reachable.
Instead in Section IV, we illustrate how to compute the forward reachable set for trajectories starting from a border and ending in another border of the same cell. In Section V, we illustrate an iterative method that combines the two abovementioned types of set to perform a global reachability analysis. Finally, in the conclusions, we discuss future developments.
\section{Problem Formulation}
\begin{figure}[t]
\centering
\begin{subfigure}[t]{0.12\textwidth}
\centering
\includegraphics[width=1\textwidth]{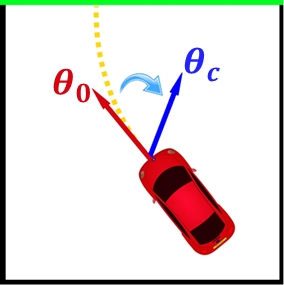}
\caption{ }
\label{fig:example_dubin_vehicle}
\end{subfigure}\quad
\begin{subfigure}[t]{0.2\textwidth}
\centering
\includegraphics[width=1\textwidth]{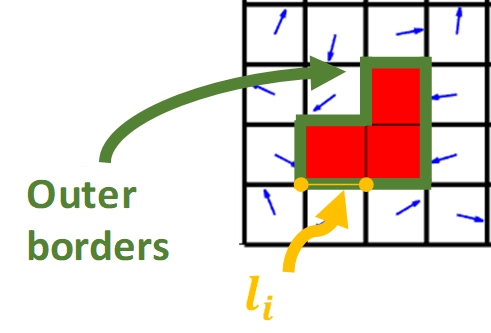}
\caption{ }
\label{fig:example_border}
\end{subfigure}
\caption{Example of vehicle moving inside a single cell (a) and global feedback plan $\pi$ with region of interest $W_{*}$ (b).}
\label{fig:example_vehicle_and_plan}
\end{figure}

Our objective is to study the backward reachability of a region of interest $W_{*}$ of the workspace $W$ for a discrete feedback plan having the following characteristics:
\begin{itemize}
\item The feedback plan is formulated in the workspace $W$ and indicates the direction with which the vehicle must travel.
\item The workspace is decomposed in a grid having cell size $d$ smaller than the vehicle's minimum turning radius $r$.
\item Bang-bang control is used to correct the error between the vehicle's heading $\theta_{v}$ and the commanded heading $\theta_{c}$ (which is constant in a cell). Therefore, the vehicle can either travel straight with speed $v$ or turn with its minimum turning radius $r=\frac{v}{\omega}$ ($\omega$ is the angular speed). As shown in \cref{fig:example_dubin_vehicle}, the vehicle turns in the direction of the minimum angular distance.
\end{itemize}

If the initial angle $\theta_{0}$ is equal to the commanded angle $\theta_{c}$, then the vehicle will just travel straight. Instead, if $\theta_{0}\neq\theta_{c}$, the vehicle will first turn and then travel straight once it reaches the commanded heading $\theta_{c}$. Nevertheless, since $d<r$ the vehicle may not converge to the desired angle before leaving the cell. In this last case, the exit angle will be different from $\theta_{c}$ and the path consists of a single turn. In summary, three types of paths are possible: 
\begin{itemize}
\item straight line (S path);
\item single turn followed by a straight line (CS path);
\item single turn (C path);
\end{itemize}
where C stands for a circular portion, which can correspond to either a right or left turn. Therefore, the set of possible paths is: $\left\lbrace S,R,L,RS,LS\right\rbrace$. 

Given a region of interest $W_{*}$, (red cells in \cref{fig:example_border}) and a feedback plan $\pi$ (blue arrows in \cref{fig:example_border}) the \emph{Backward Reachable Set} of $W_{*}$ is the set of all the configurations from which $W_{*}$ is reachable. In particular, a configuration $q_{0}$ is in the backward reachable set of $W_{*}$, if starting from $q_{0}$ and following the feedback plan $\pi$, there exists a time $t_{*}$ in which the vehicle's configuration $q_{*}$ in the set $W\times\mathbb{S}^{1}$ (i.e. inside the cell occupied by $W_{*}$). What stated above can be summarized as follows:
\begin{equation}
\begin{aligned}
B\left(W_{*},r,\pi\right) =& \{q_{0}\in C=W\times\mathbb{S}^{1} \vert \exists t_{*} \in \left[ 0,\infty\right)\\ \text{such that }
&q\left(t_{*}\right)\in C_{\text{*}}= W_{*}\times\mathbb{S}^{1} \}
\label{eq:def_backward_reach_set_tot}
\end{aligned}
\end{equation}

The region of interest $W_{*}$ can be represented with its outer borders as illustrated in \cref{fig:example_border}. The generic outer border is $l_{i}$, while the set of all the outer borders is $L=\bigcup\limits_{i=1}^{n} l_{i}$. We can reformulate the backward reachable set by observing that $W_{*}$ is reachable from $q_{0}$ if the trajectory starting from $q_{0}$ intersects $L$. Exploiting this observation, the backward reachable set becomes:
\begin{equation}
\begin{aligned}
B\left(L,r,\pi\right)=\{q_{0}\in C=W\times\mathbb{S}^{1}\vert \exists t_{*} \in \left[ 0,\infty\right) \\ \text{such that } \left(x\left(t_{*}\right),y\left(t_{*}\right)\right)\in L=\bigcup\limits_{i=1}^{n} l_{i}\}
\label{eq:def_backward_reach_set_side}
\end{aligned}
\end{equation} 
This second formulation allows us to encode $W_{*}$ and $C_{*}$, which are respectively 2-D and 3-D subspaces, with the union of respectively 1-D sets $l_{i}$ (outer borders) and 2-D regions $l_{i}\times\mathbb{S}^{1}$ (coordinate and heading angles at the borders).

\section{Cellular Backward Reachability}
\begin{figure}[t]
\begin{subfigure}[t]{0.20\textwidth}
\includegraphics[width=1\textwidth]{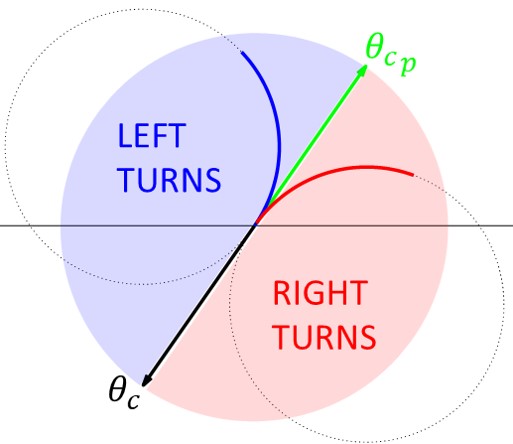}
\caption{ }\label{fig:rule_turns}
\end{subfigure}\hfill
\begin{subfigure}[t]{0.27\textwidth}
\includegraphics[width=1\textwidth]{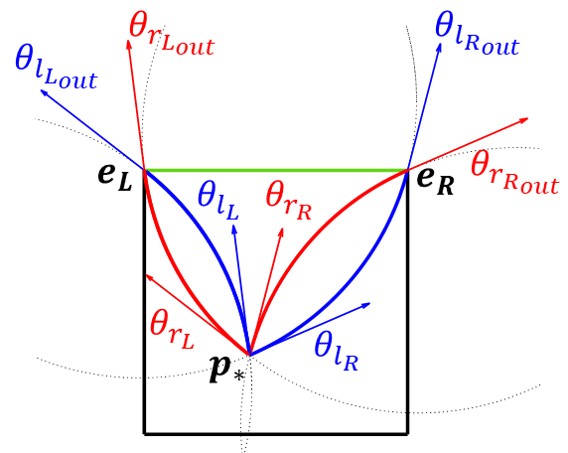}
\caption{ }\label{fig:reach_top_border}
\end{subfigure}
\caption{Regions of left and right turns (a). Maximum and minimum turns for trajectories starting from the considered point $p_{*}$ and ending in the top border's edges (b).}\label{fig:turn_rules and_reach_top_border}
\end{figure}

In this section, we describe how to compute the backward reachable set of a border $l_{i}$ in a point $p_{*}=(x_{*},y_{*})$ inside a cell. We start by observing (\cref{fig:rule_turns}) that the commanded direction $\theta_{c}$ and the quantity $\theta_{c_{p}}=\theta_{c}+\pi$ delimit two regions: \emph{left turns} (LT) and \emph{right turns} (RT). The vehicle's orientation $\theta$ can be either in one of the abovementioned regions or equal to $\theta_{c}$. Consequentially, the control law can be formulated as follows:
\begin{equation}
u\left(\theta,\theta_{c}\right)=\begin{cases}
                        \omega \quad \text{if $\theta\in \textit{LT}$} \\
                        0 \quad \text{if $\theta\equiv \theta_{c}$}\\
                        -\omega \quad \text{if $\theta\in \textit{RT}$}
                    \end{cases}
\label{eq:control_law_dubin2}
\end{equation}
In order to establish the direction of the turn, $\theta$ must be compared with $\theta_{c}$ and $\theta_{c_{p}}$. In our formulation, we assume that angles less than $\frac{\pi}{2}$ and greater then $-\frac{\pi}{2}$ are compared wrapping them in the interval $[-\pi,\pi]$, while angles outside this range are compared wrapping them in the interval $[0,2\pi]$. 

We illustrate the formulation only for the case of \emph{top} border backward reachability (\cref{fig:reach_top_border}), because for other borders we can exploit the same principles.

The backward reachable set $B(l_{\text{top}},r,\theta_{c})$ ($l_{\text{top}}$ means \emph{top} border) for a point $p_{*}=\left(x_{*},y_{*}\right)$ is the set of angles from which the top border is reachable. Such a set is continuous and delimited by a minimum and a maximum provided by the two piece-wise functions $\Theta_{\text{min}}\left(x,y,\theta_{c}\right)$ and $\Theta_{\text{max}}\left(x,y,\theta_{c}\right)$ \footnote{We omit the dependence from the cell size $d$ and the minimum turning radius $r$.}. Our objective in this section is to illustrate the geometric principles that are used to derive the functions $\Theta_{\text{min}}\left(x,y,\theta_{c}\right)$ and $\Theta_{\text{max}}\left(x,y,\theta_{c}\right)$. 

From \cref{fig:reach_top_border} we can infer that minimum and maximum angles can be found by considering the top border's edges. There are two circles having radius $r$ passing through the point $p_{*}$ and the edge $e_{*}$. Each circle corresponds to either a left turn or a right turn and the centers of the two circles are obtained as follows:
\begin{equation}
\begin{aligned}
q&=\sqrt{\left(x_{*}-x_{e_{*}}\right)^{2}+\left(y_{*}-y_{e_{*}}\right)^{2}}\\
x_{c_{1,2}}&=\frac{x_{*}+x_{e_{*}}}{2}\pm \frac{y_{e_{*}}-y_{*}}{q}\sqrt{r^{2}-{\left(\frac{q}{2}\right)^{2}}}\\
y_{c_{1,2}}&=\frac{y_{*}+y_{e_{*}}}{2}\pm \frac{x_{*}-x_{e_{*}}}{q}\sqrt{r^{2}-{\left(\frac{q}{2}\right)^{2}}}
\end{aligned}
\label{eq:circle_2_points}
\end{equation}

\begin{figure}[t]
\begin{minipage}{0.15\textwidth}
\begin{subfigure}[t]{1\textwidth}
        \includegraphics[width=1\textwidth]{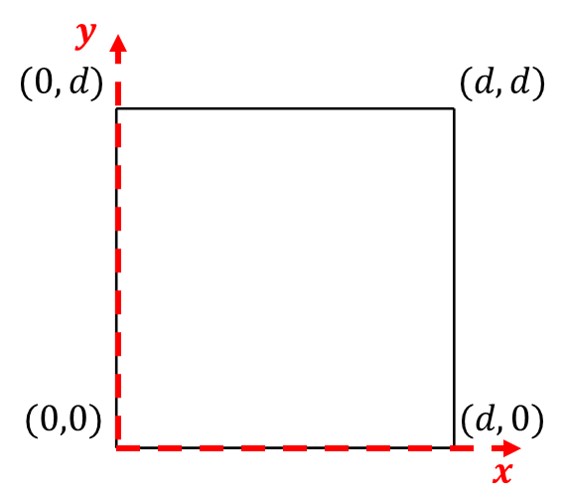}
    \caption{ }\label{fig:local_frame}
\end{subfigure}
\end{minipage}
\begin{minipage}{0.35\textwidth}
\begin{subfigure}[b]{0.49\textwidth}\centering
\includegraphics[width=1\textwidth]{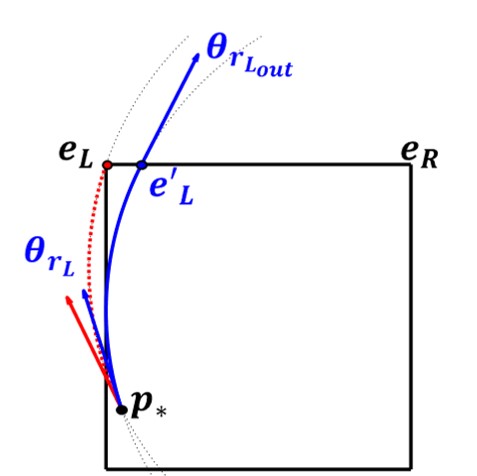}
\caption{ }\label{fig:ex_tang_left}
\end{subfigure}
\begin{subfigure}[b]{0.49\textwidth}\centering
\includegraphics[width=1\textwidth]{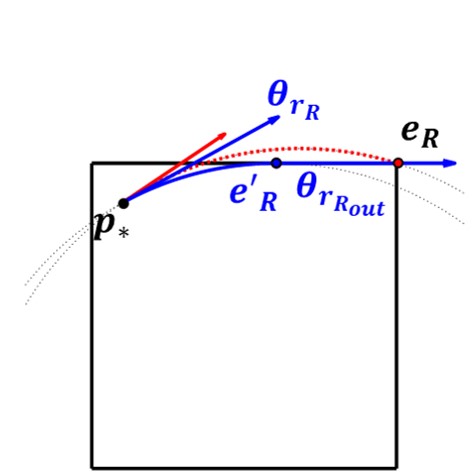}
\caption{ }\label{fig:ex_tang_top_right}
\end{subfigure}\\
\begin{subfigure}[b]{0.49\textwidth}\centering
\includegraphics[width=1\textwidth]{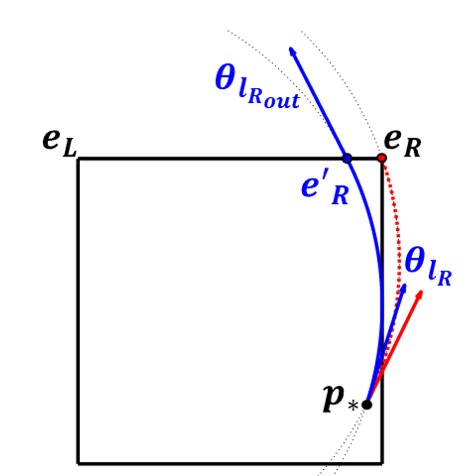}
\caption{ }\label{fig:ex_tang_right}
\end{subfigure}
\begin{subfigure}[b]{0.49\textwidth}\centering
\includegraphics[width=1\textwidth]{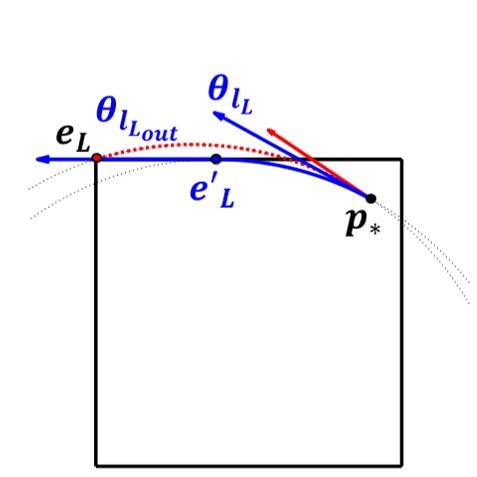}
\caption{ }\label{fig:ex_tang_top_left}
\end{subfigure}
\end{minipage}
\caption{Local reference frame (a). Right turn tangential to the left border (b), right turn tangential to the top border (c), left turn tangential to the right border (d), left turn tangential to the top border (e).}\label{fig:examples_tang}
\end{figure}

In our formulation, we use the local reference frame illustrated in \cref{fig:local_frame}.

With the coordinates $x_{c},y_{c}$ we can find the initial angles depicted in \cref{fig:reach_top_border}. In particular, for left turns we can find the direction from the center $(x_{c_{l_{\#}}},y_{c_{l_{\#}}})$ ($\#$ is the type of edge, which is either L or R) towards the considered point $p_{*}$ and then rotate it counterclockwise of 90 degrees:
\begin{equation}
\theta_{l_{\#}}= \atantwo(y_{*}-y_{c_{l_{\#}}},x_{*}-x_{c_{l_{\#}}})+\frac{\pi}{2}
\label{eq:left_turn_angle}
\end{equation}
For right turns, the rotation must be 90 degrees clockwise:
\begin{equation}
\theta_{r_{\#}}= \atantwo(y_{*}-y_{c_{r_{\#}}},x_{*}-x_{c_{r_{\#}}})-\frac{\pi}{2}
\label{eq:right_turn_angle}
\end{equation}

The set identified by $\Theta_{\text{min}}\left(x,y,\theta_{c}\right)$ and $\Theta_{\text{max}}\left(x,y,\theta_{c}\right)$ must contain only trajectories that are entirely contained in the cell. This requirement implies that in some location the minimum or maximum left/right turn is not given by a circle intersecting an edge. In particular, it might be necessary to find the circle tangential to one of the side borders or the top border. In \cref{fig:examples_tang}, there are examples showing when it is necessary to use the circles tangential to the borders. In those examples, continuous blue lines are tangential turns, while dotted red lines are turns given by circles intersecting the edges.

In order to establish if the trajectory exits from one of the side borders or a point internal to the top border we must test the exit angles. In particular, we have four cases:
\begin{itemize}
\item if $\theta_{r_{L_{out}}}<\frac{\pi}{2}$ (\cref{fig:ex_tang_left}), find right turn tangential to left border;
\item if $\theta_{r_{R_{out}}}<0$ (\cref{fig:ex_tang_top_right}), find right turn tangential to top border;
\item if $\theta_{l_{R_{\text{out}}}}>\frac{\pi}{2}$ (\cref{fig:ex_tang_right}), find left turn tangential to right border;
\item if $\theta_{l_{L_{out}}}>\pi$ (\cref{fig:ex_tang_top_left}), find left turn tangential to top border;
\end{itemize}

The circle tangential to a border can be easily obtained by imposing the intersection in the point $p_{*}$ and that its distance from the border is equal to $r$. Imposing these two conditions, two circles are obtained; thus, we must make sure to pick the correct solution. For instance in the case of left turn tangential to right border we have:
\begin{equation}
\begin{aligned}
x_{c_{l_{\text{tan}_{R}}}}&=d-r\\
y_{c_{l_{\text{tan}_{R}}}}&=y_{*}+\sqrt{r^{2}-(x_{c_{l_{\text{tan}_{R}}}}-x_{*})^{2}}
\label{eq:tan_right_border}
\end{aligned}
\end{equation}

The functions $\Theta_{min}\left(x,y,\theta_{c}\right)$ and $\Theta_{max}\left(x,y,\theta_{c}\right)$ are piece-wise and their conditions are obtained by comparing $\theta_{*_{\#}}$ and $\theta_{*_{\#_{out}}}$ with either $\theta_{c}$ or $\theta_{c_{p}}$.  In the following sections, we illustrate how to obtain the two functions making the distinction between the two types of commands: $\theta_{c}\in [0,\pi]$ (forward command) and $\theta_{c}\in(\pi,2\pi)$ (backward command).

\subsection{Top Border Backward Reachable set: Forward Command}
\begin{figure}[t]
\centering
\begin{subfigure}[t]{0.16\textwidth}\centering
\includegraphics[width=1\textwidth]{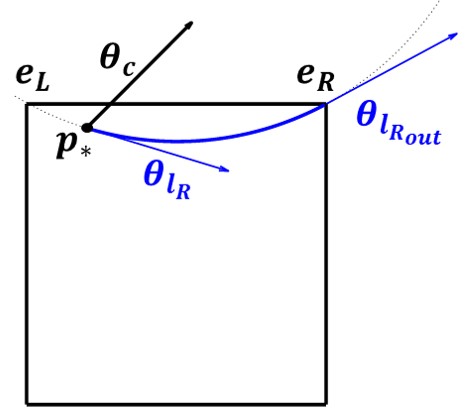}
\caption{ }\label{fig:normal_min_left}
\end{subfigure}\hfill\begin{subfigure}[t]{0.16\textwidth}\centering
\includegraphics[width=1\textwidth]{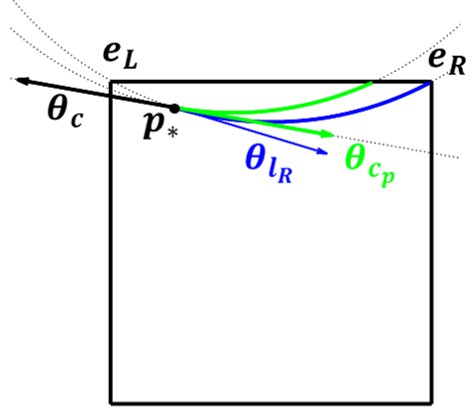}
\caption{ }\label{fig:limited_min_left}
\end{subfigure}\hfill
\begin{subfigure}[t]{0.16\textwidth}\centering
\includegraphics[width=1\textwidth]{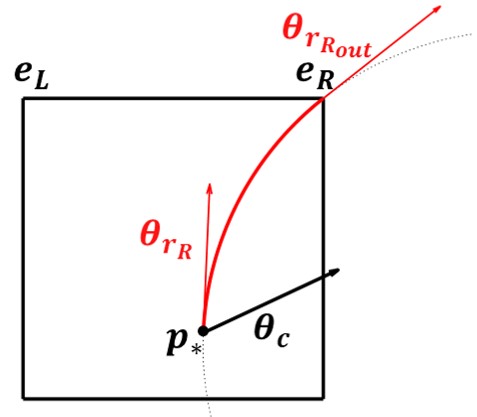}
\caption{ }\label{fig:normal_min_right}
\end{subfigure}\hfill\\
\begin{subfigure}[t]{0.16\textwidth}\centering
\includegraphics[width=1\textwidth]{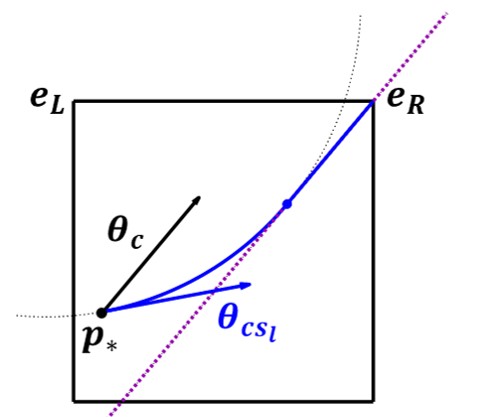}
\caption{ }\label{fig:cs_l_R}
\end{subfigure}
\begin{subfigure}[t]{0.16\textwidth}\centering
\includegraphics[width=1\textwidth]{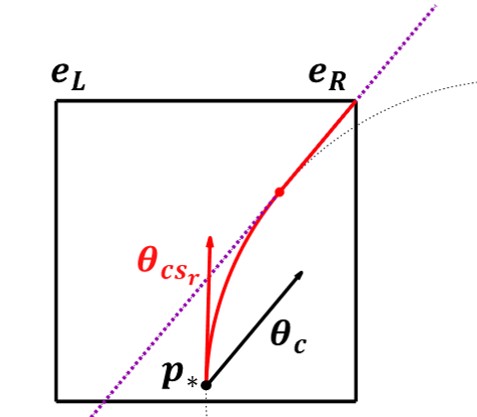}
\caption{ }\label{fig:cs_r_R}
\end{subfigure}
\caption{Cases for minimum: left turn (a), \emph{limited} left turn (b), right turn (c), LS path (d),  RS path (e). }\label{fig:min_turns_back_cell}
\end{figure}

In the case of a forward command, the type of curve for minimum and maximum heading angle is determined by comparing $\theta_{c}$ with both the initial and exit angles. 
If $\theta_{c}>\theta_{l_{R}}$ and $\theta_{c}\geq\theta_{l_{R_{\text{out}}}}$, then the minimum is given by a left turn (\cref{fig:normal_min_left,fig:limited_min_left}) and it is equal to $\max(\theta_{c_{p}},\theta_{l_{R}})$ (if $\theta_{c_{p}}>\theta_{l_{R}}$, $\theta_{l_{R}}$ is in the RT region, therefore a left turn cannot start with it). Instead, if $\theta_{c}<\theta_{r_{R}}$ and $\theta_{c}\leq\theta_{r_{R_{\text{out}}}}$, then the minimum is given by a right turn (\cref{fig:normal_min_right}) and it is equal to $\theta_{r_{R}}$. If none of the abovementioned conditions is true, then the minimum will be given by either an S or a CS path. In this case the minimum is obtained using the geometric construction depicted in \cref{fig:cs_l_R,fig:cs_r_R}.
The straight line (CS border) intersecting the right edge $e_{R}$ and having equation:
\begin{equation}
y=mx+q,\qquad m=\tan(\theta_{c}),\qquad q=(1-m)d
\label{eq:cs_case_right}
\end{equation} splits the cell in three regions. A point $p_{*}$ can be above the line (A region), on the line (L region), or below the line (B region):  
\begin{enumerate}
\item $y>mx+q$ (\emph{A region})
\item $y=mx+q$ (\emph{L region})
\item $y<mx+q$ (\emph{B region})
\end{enumerate}
For locations lying on the border, the minimum is $\theta_{c}$ and it is given by a straight line. Instead for positions that are in the regions above and below the CS border, the minimum is given by a left and a right CS path, respectively. The exit angle is equal to the commanded heading angle $\theta_{c}$, while the minimum is computed using the circle that is tangential to the CS border and that intersects the point $p_{*}$. If $p_{*}\in A$, the circle tangential to the CS border corresponds to a left turn. The center of the circle and the corresponding initial angle $\theta_{\text{cs}_{l}}$ are computed as follows:
\begin{equation}
\begin{aligned}
q_{p}&= q+\frac{r}{\vert\sin(\frac{\pi}{2}-\theta_{c})\vert}\\
\beta&=\frac{2mq_{p}-2x_{*}-2my_{*}}{1+m^{2}}\\
\gamma&=\frac{x_{*}^{2}+y_{*}^{2}+q_{p}^{2}-2y_{*}q_{p}-r^{2}}{1+m^{2}}\\
x_{\text{cs}_{l}}&=\frac{-\beta+\sqrt{\beta^{2}-4\gamma}}{2}\\
y_{\text{cs}_{l}}&=mx_{\text{cs}_{l}}+q_{p}\\
\theta_{\text{cs}_{l}}&=\atantwo(x_{*}-x_{\text{cs}_{l}},y_{\text{cs}_{l}}-y_{*})+\frac{\pi}{2}
\end{aligned}
\label{eq:tang_CS_border_up_min}
\end{equation}
where $q$ and $m$ are the ones computed in \cref{eq:cs_case_right}. When  $p_{*}\in B$, the circle tangential to the CS border corresponds to a right turn. In this case, the equations for the circle and the initial angle $\theta_{\text{cs}_{r}}$ are the same as in \cref{eq:tang_CS_border_up_min} except for $q_{p}$ and the angle $\theta_{\text{cs}_{r}}$, which are computed as follows:
\begin{equation}
\begin{aligned}
q_{p}&= q-\frac{r}{\vert\sin(\frac{\pi}{2}-\theta_{c})\vert}\\
\theta_{\text{cs}_{r}}&=\atantwo(x_{\text{cs}_{r}}-x_{*},y_{*}-y_{\text{cs}_{r}})-\frac{\pi}{2}
\end{aligned}
\label{eq:tang_CS_border_down_min}
\end{equation}

\begin{figure}[t]
\centering
\begin{subfigure}[t]{0.16\textwidth}\centering
\includegraphics[width=1\textwidth]{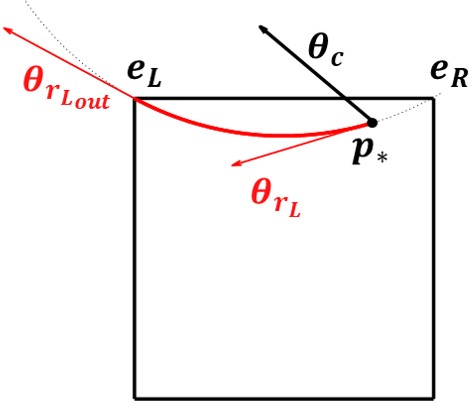}
\caption{ }\label{fig:normal_max_right}
\end{subfigure}\hfill\begin{subfigure}[t]{0.16\textwidth}\centering
\includegraphics[width=1\textwidth]{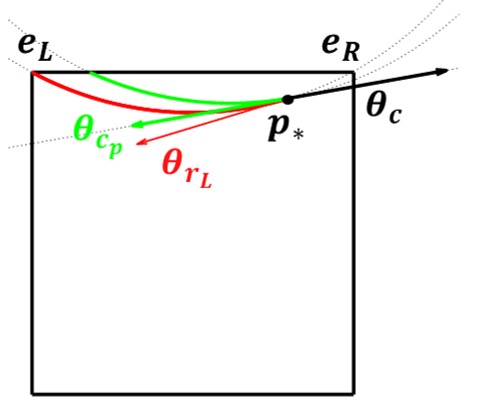}
\caption{ }\label{fig:limited_max_right}
\end{subfigure}\hfill
\begin{subfigure}[t]{0.16\textwidth}\centering
\includegraphics[width=1\textwidth]{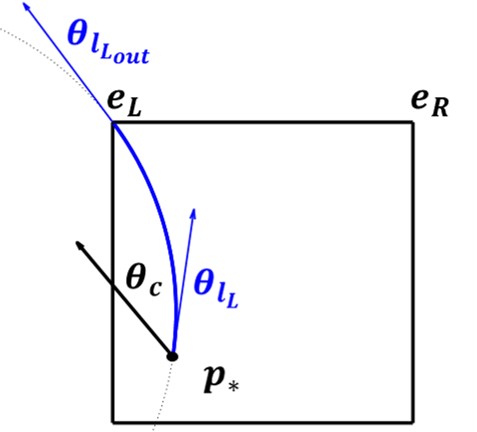}
\caption{ }\label{fig:normal_max_left}
\end{subfigure}\hfill\\
\begin{subfigure}[t]{0.16\textwidth}\centering
\includegraphics[width=1\textwidth]{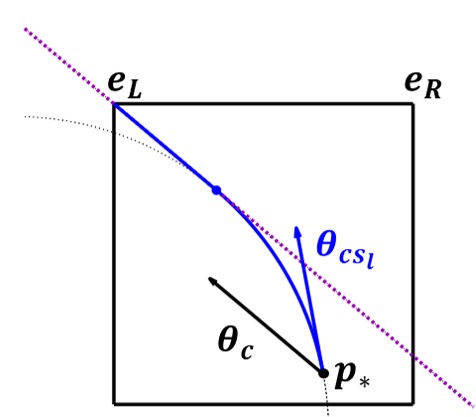}
\caption{ }\label{fig:cs_l_L}
\end{subfigure}
\begin{subfigure}[t]{0.16\textwidth}\centering
\includegraphics[width=1\textwidth]{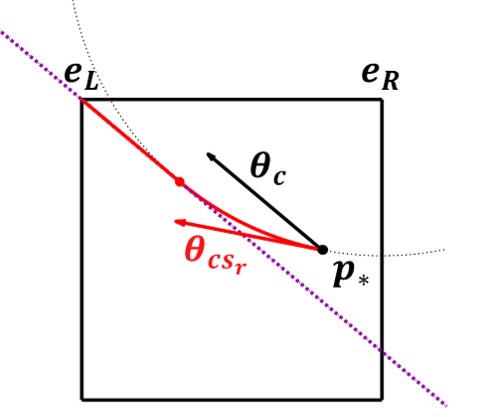}
\caption{ }\label{fig:cs_r_L}
\end{subfigure}
\caption{Cases for maximum: right turn (a), \emph{limited} right turn (b), left turn (c), LS path (d), RS path (e).}\label{fig:max_turns_back_cell}
\end{figure}

\begin{figure}[t]
\centering
\includegraphics[width=0.33\textwidth]{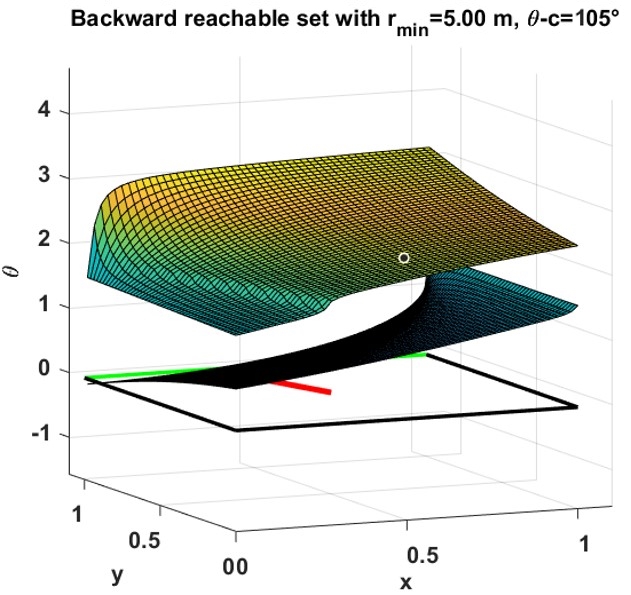}
\caption{Example of backward reachable set for forward command.}\label{fig:TOP_back_set_F}
\end{figure}

The formulation for $\Theta_{max}(x,y,\theta_{c})$ is based on the same considerations used for the minimum. In this case, we must deal with the left side of the cell as illustrated in \cref{fig:max_turns_back_cell}. 
If $\theta_{c}<\theta_{r_{L}}$ and $\theta_{c}\leq\theta_{r_{L_{\text{out}}}}$, then the maximum is given by a right turn (\cref{fig:normal_max_right,fig:limited_max_right}) and it is equal to $\min(\theta_{c_{p}},\theta_{r_{L}})$ (if $\theta_{c_{p}}<\theta_{r_{L}}$, $\theta_{r_{L}}$ is in the LT region, therefore a right turn cannot start with it). When $\theta_{c}>\theta_{l_{L}}$ and $\theta_{c}\geq\theta_{l_{L_{\text{out}}}}$, then the maximum is given by a left turn (\cref{fig:normal_min_right}) and it is equal to $\theta_{l_{L}}$. If none of the above conditions is true, then the maximum is given by either a S or a CS path. In this case we use the equations for the CS border used before with the following differences:
\begin{itemize}
\item $q=d$ (the CS border must intersect the left edge $e_{L}$);
\item a circle in \textit{region A} corresponds to a right turn, while a circle in \textit{region B} corresponds to a left turn as shown in \cref{fig:cs_r_L,fig:cs_l_L}.
\end{itemize}
In \cref{fig:TOP_back_set_F} there in an examples of cellular backward reachable set (for the top border) in the case of forward command. 

\subsection{Top Border Backward Reachable set: Backward Command}
In this section, we study the backward reachability for $\theta_{c}\in(\pi,2\pi)$. The first difference with the previous case is that only C paths are admissible. To determine the direction of the minimum and maximum turns we must compare angles with $\theta_{c_{p}}$.
\begin{figure}[t]
	\centering
	\begin{subfigure}[b]{0.14\textwidth}
	\centering	\includegraphics[width=1\textwidth]{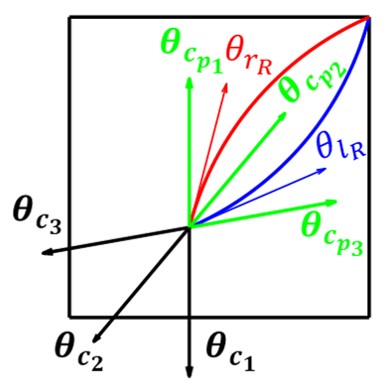}
		\caption{ }\label{fig:region_min_back}
	\end{subfigure}\qquad
	\begin{subfigure}[b]{0.14\textwidth}
		\centering	\includegraphics[width=1\textwidth]{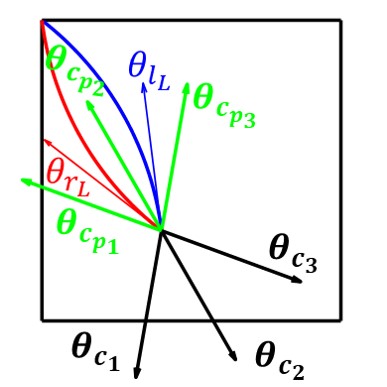}
		\caption{ }\label{fig:region_max_back}
	\end{subfigure}
	\caption{Angle regions for minimum (a) and maximum (b) in the case of backward command.}\label{fig:region_back_cellular}
\end{figure}

Referring to \cref{fig:region_back_cellular}, we can see that for the definition of the minimum and the maximum we can identify three angle intervals. 
For the minimum, the first interval is $\theta_{c_{p_{1}}}\geq\theta_{r_{R}}$ and the minimum is given by a right turn with starting angle $\theta_{r_{R}}$. Instead for the maximum, in the first interval we have $\theta_{c_{p_{1}}}\geq\theta_{r_{L}}$ and the maximum is $\theta_{r_{L}}$.

In the second interval, the condition for the minimum is $\theta_{l_{R}}<\theta_{c_{p_{2}}}<\theta_{r_{R}}$ and the minimum is given by a left turn with starting angle $\theta_{c_{p_{2}}}$ (a right turn would intersects the right border rather than the top border). While for the maximum we have $\theta_{l_{L}}<\theta_{c_{p_{2}}}<\theta_{r_{L}}$ and the maximum is $\theta_{c_{p_{2}}}$.

Finally, in the last region the condition for the minimum is $\theta_{c_{p_{3}}}\leq\theta_{l_{R}}$ and the minimum is $\theta_{l_{R}}$ (the angle cannot be smaller, otherwise the circle would intersect the right border); while the condition for the maximum is $\theta_{c_{p_{3}}}\leq\theta_{l_{L}}$ and the maximum is $\theta_{l_{L}}$. 

\begin{figure}[t]
    \centering
    \begin{subfigure}[b]{0.22\textwidth}
    \centering
        \includegraphics[width=1\textwidth]{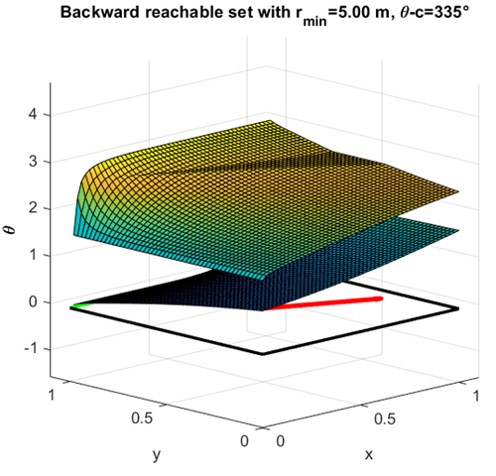}
    \caption{ }\label{fig:TOP_back_set_B}
	\end{subfigure} \qquad
	\begin{subfigure}[b]{0.2\textwidth}
    \centering
        \includegraphics[width=1\textwidth]{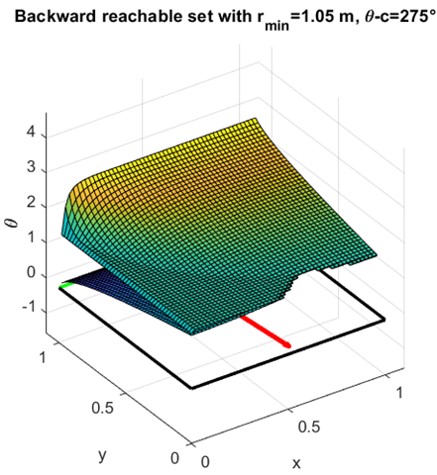}
    \caption{ }
    \label{fig:TOP_back_set_BE}
    \end{subfigure}
    \caption{Examples of backward reachable set for backward command.}
\end{figure}

With backward commands the top border might not be reachable from some location. This happens when the minimum turning radius is close to the cell size. In particular, if the minimum is associated to a left turn and $\theta_{\text{min}}>\theta_{l_{L}}$, then such a turn does not lead to the top border. In fact, any left turn with initial angle greater than $\theta_{l_{L}}$ exits the cell from the left border. For the maximum there is a similar problem. There is no solution if the maximum is given by a right turn with $\theta_{\text{max}}<\theta_{r_{R}}$. The complete condition for the existence of a solution is the following:
\begin{equation}
\begin{aligned}
B_{x_{*},y_{*}}(l_{\text{top}},r_{\text{min}},\theta_{c})=\emptyset\Leftrightarrow(u(\theta_{\text{min}},\theta_{c})=\omega\wedge\theta_{\text{min}}>\theta_{l_{L}}))\\ \vee(u(\theta_{\text{max}},\theta_{c})=-\omega\wedge\theta_{\text{max}}<\theta_{r_{R}})
\end{aligned}
\label{eq:condition_no_reachable}
\end{equation}
In \cref{fig:TOP_back_set_B} there is an  example in which for every point there exists a solution, while \cref{fig:TOP_back_set_BE} illustrates an example where the top border is not reachable from some locations.

\subsection{Border to Border Backward Reachability}
In our method, we propagate sets from one border to another, therefore we use the univariate functions $\Theta^{\#}_{*_{\text{min}}}(x,\theta_{c})$ and $\Theta^{\#}_{*_{\text{max}}}(x,\theta_{c})$, where $\#$ is the arrival border and $*$ is the starting border. For instance, $\Theta^{T}_{R_{\text{min}}}(x,\theta_{c})$ and $\Theta^{T}_{R_{\text{max}}}(x,\theta_{c})$ provide the top border backward reachable set in the right border ($x$ is the coordinate at the border according to the local reference frame in \cref{fig:local_frame}) and they are obtained from $\Theta_{\text{min}}(x,y,\theta_{c})$ and $\Theta_{\text{max}}(x,y,\theta_{c})$ imposing $x=d$.

\section{Border to Border Forward Reachability}
In the previous sections, the functions $\Theta^{A}_{B_{\text{min}}}(x,\theta_{c})$ and $\Theta^{A}_{B_{\text{max}}}(x,\theta_{c})$ provide the set of all the angles that a trajectory starting from a point of border B can have in order to reach border A. In this case, the arrival position and heading angle are not considered. Instead, in this section, we deal with the opposite problem, which consists in deriving the piece-wise functions $\Phi^{A}_{B_{\text{min}}}(x,\theta_{c})$ and $\Phi^{A}_{B_{\text{max}}}(x,\theta_{c})$ that provide minimum and maximum arrival angles in a point of border A for trajectories starting from border B. Therefore, this two functions define the \emph{forward} reachable set of the border A at the border B. 

The process is illustrated for one border because the functions for other borders can be obtained using the same geometric principles. In particular, we illustrate the cases in which the bottom border is the starting border and the top and right borders are the ending borders. The case of left border is dual to the case of the right border, therefore we do not provide a detailed explanation. 

In the description, for the top border we distinguish between: $\theta_{c}\in[0,\pi]$ (forward command) and $\theta_{c}\in(\pi,2\pi)$ (backward command), while for the right border we distinguish between  $\theta_{c}\in[-\frac{\pi}{2},\frac{\pi}{2}]$ (right command) and $\theta_{c}\in(\frac{\pi}{2},\frac{3}{2}\pi)$ (left command).

\subsection{Bottom Border Forward Reachable Set at Top Border: Forward Command}
\begin{figure}[t]
\begin{subfigure}[t]{0.15\textwidth}\centering
\includegraphics[width=1\textwidth]{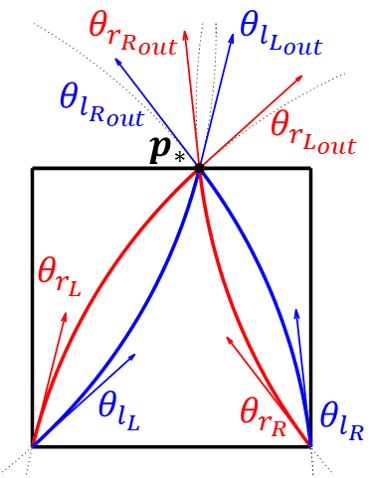}
\caption{ }\label{fig:reach_top_border_forward}
\end{subfigure}\hfill
\begin{subfigure}[t]{0.15\textwidth}\centering
\includegraphics[width=1\textwidth]{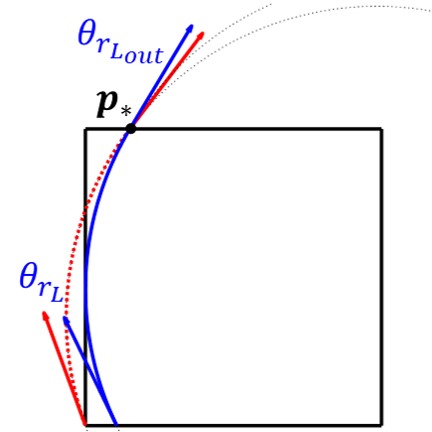}
\caption{ }\label{fig:example_tang_left_forward}
\end{subfigure}\hfill
\begin{subfigure}[t]{0.15\textwidth}\centering
\includegraphics[width=1\textwidth]{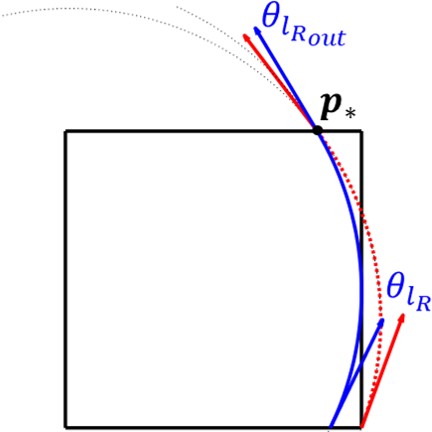}
\caption{ }\label{fig:example_tang_right_forward}
\end{subfigure}
\caption{Angles for forward reachability (a). Right turn tangential to left border (b). Left turn tangential to right border (c).}\label{fig:intro_forward}
\end{figure}

We are considering the forward reachability for the bottom border; therefore, we must consider turns that start from the edges of the bottom border. The ending point $p_{*}$ is located in the top border. Like the case of backward reachability, \cref{eq:circle_2_points} is used to find the circles intersecting the edges and the considered point $p_{*}$. In this case, with the local reference frame in \cref{fig:local_frame}, $y_{*}$ is always equal to $d$, while the edges of the bottom border are $e_{L}=(0,0)$ and $e_{R}=(d,0)$. 
Also for the forward reachable set, only trajectories that are entirely contained in the cell must be considered. In particular, for trajectories starting from the bottom border and ending at the top border there are two cases in which the tangential circle must be computed: $\theta_{r_{L}}>\frac{\pi}{2}$ (\cref{fig:example_tang_left_forward}) and $\theta_{l_{R}}<\frac{\pi}{2}$ (\cref{fig:example_tang_right_forward}).

For forward commands, we must evaluate the exit angles and there are two possible scenarios: $\theta_{r_{R_{\text{out}}}}<\theta_{l_{L_{\text{out}}}}$ and $\theta_{r_{R_{\text{out}}}}\geq\theta_{l_{L_{\text{out}}}}$

\begin{figure}[]
\begin{subfigure}[t]{0.15\textwidth}\centering
\includegraphics[width=1\textwidth]{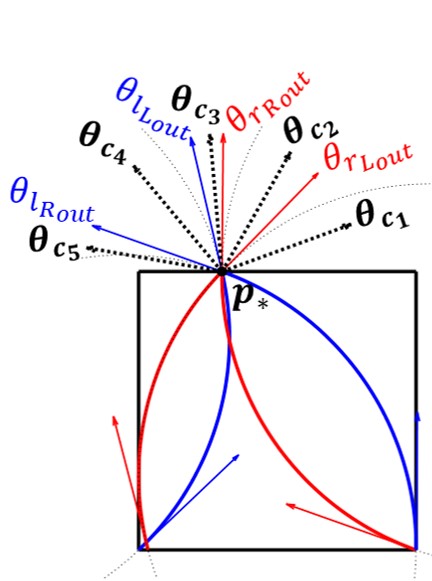}
\caption{ }\label{fig:forward_top_case_A}
\end{subfigure}\hfill
\begin{subfigure}[t]{0.15\textwidth}\centering
\includegraphics[width=1\textwidth]{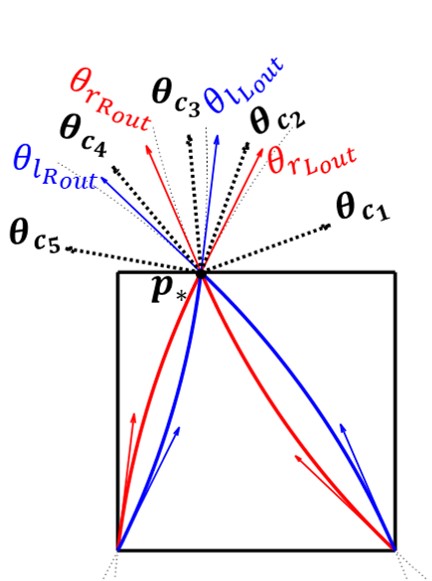}
\caption{ }\label{fig:foward_top_case_B}
\end{subfigure}\hfill
\begin{subfigure}[t]{0.15\textwidth}\centering
\includegraphics[width=1\textwidth]{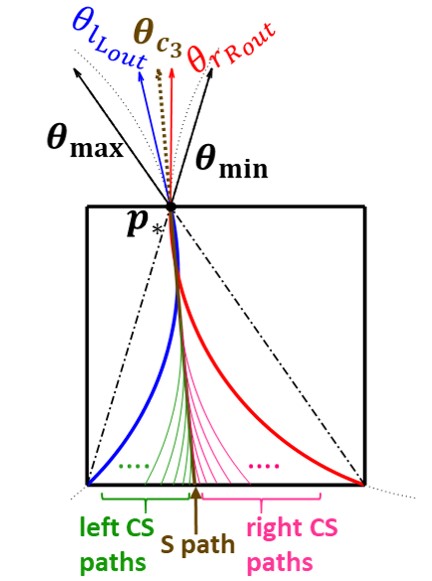}
\caption{ }\label{fig:forward_top_CS_sets}
\end{subfigure}
\caption{Angle intervals for: $\theta_{r_{R_{\text{out}}}}<\theta_{l_{L_{\text{out}}}}$ (a), $\theta_{r_{R_{\text{out}}}}\geq\theta_{l_{L_{\text{out}}}}$ (b). Example of forward reachable set at the top border consisting in a singleton (c). }\label{fig:foward_top_case}
\end{figure}

We start considering the first one, which is illustrated in \cref{fig:forward_top_case_A}. We can identify five angle intervals. For the case of commanded heading angle $\theta_{c_{1}}\leq\theta_{r_{L_{\text{out}}}}$, both minimum and maximum are given by a right turn and they are respectively $\theta_{r_{L_{\text{out}}}}$ and $\theta_{r_{R_{\text{out}}}}$. In fact, all the trajectories starting from the bottom border attempt to reach $\theta_{c_{1}}$, but none of them is able to reach it (the minimum reachable angle is $\theta_{r_{L_{\text{out}}}}$). 

In the second interval, the commanded heading angle is $\theta_{r_{L_{\text{out}}}}<\theta_{c_{2}}\leq\theta_{r_{R_{\text{out}}}}$. For angles in this interval the maximum is still $\theta_{r_{R_{\text{out}}}}$, instead the minimum is $\theta_{c_{2}}$. Therefore, the maximum is given by an R path, while the minimum by an RS path.

In the interval given by the condition $\theta_{r_{R_{\text{out}}}}<\theta_{c_{3}}<\theta_{l_{L_{\text{out}}}}$, neither left nor right turns are possible. In this case, minimum and maximum are exactly $\theta_{c_{3}}$. Therefore, such a forward configuration is the end configuration for an S or a CS path. In \cref{fig:forward_top_CS_sets}, the angles $\theta_{\text{min}}$ and $\theta_{\text{max}}$ are obtained from the slopes of the straight lines passing through the considered point and respectively left and right edges of the bottom border. Since $\theta_{\text{min}}<\theta_{c_{3}}<\theta_{\text{max}}$, there always exist an S path, a set of RS paths and a set of LS paths. The elements of these sets are obtained by varying the position in which the circle intersects the bottom border. 

The fourth and fifth intervals are given by the condition $\theta_{l_{L_{\text{out}}}}<\theta_{c_{4}}< \theta_{l_{R_{\text{out}}}}$ and $\theta_{c_{5}}\geq\theta_{l_{R_{\text{out}}}}$. These last two cases are dual to the first two cases. For both ranges, the minimum is given by a left turn and it is $\theta_{l_{L_{\text{out}}}}$. Instead the maximum is different for the two cases. In the first one, it is $\theta_{c_{4}}$ and it is given by LS paths, while in the second case it is $\theta_{l_{R_{\text{out}}}}$ and it corresponds to a left turn.

When $\theta_{r_{R_{\text{out}}}}\geq\theta_{l_{L_{\text{out}}}}$ there are still five intervals as shown in \cref{fig:foward_top_case_B}. For the following four intervals:
\begin{itemize}
\item $\theta_{c_{1}}\leq\theta_{r_{L_{\text{out}}}}$
\item $\theta_{r_{L_{\text{out}}}}<\theta_{c_{2}}<\theta_{l_{L_{\text{out}}}}$
\item $\theta_{r_{R_{\text{out}}}}<\theta_{c_{4}}<\theta_{l_{R_{\text{out}}}}$
\item $\theta_{l_{R_{\text{out}}}}<\theta_{c_{5}}$
\end{itemize} 
the rules derived previously for the angles $\theta_{c_{1}}$, $\theta_{c_{2}}$, $\theta_{c_{4}}$, and  $\theta_{c_{5}}$ are still valid. Instead for $\theta_{c_{3}}$, now we have the condition $\theta_{l_{L_{\text{out}}}}<\theta_{c_{3}}\leq\theta_{r_{R_{\text{out}}}}$. In this case, we do not have a singleton anymore, instead we have a set having minimum $\theta_{l_{L_{\text{out}}}}$ given by a left turn and maximum $\theta_{c_{3}}$ given by CS paths. 

\subsection{Bottom Border Forward Reachable Set at the Top Border: Backward Command}
Here we consider the case $\theta_{c}\in(\pi,2\pi)$. The first substantial difference between the two cases is that while for forward commands the reachable set evaluated in one point is a continuous set delimited by a minimum and a maximum, for backward commands there are scenarios in which the reachable set splits in two regions.
The other difference is that rather than comparing the exit angles with the commanded heading angle $\theta_{c}$, we compare the starting angles with the angle $\theta_{c_{p}}=\theta_{c}+\pi$. Furthermore, S and CS  paths cannot exist.

\begin{figure}[t]
	\centering	\includegraphics[width=0.3\textwidth]{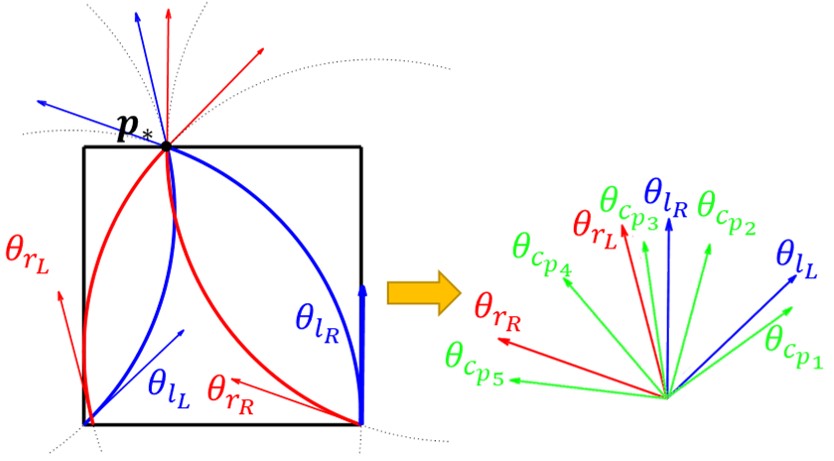}
		\caption{Angle regions for $\theta_{l_{R}}<\theta_{r_{L}}$.}\label{fig:reach_top_border_forward_back_cmd_A}
\end{figure}

To derive the functions in this case, we make the following distinction: $\theta_{l_{R}}<\theta_{r_{L}}$ and $\theta_{l_{R}}\geq\theta_{r_{L}}$. We start considering the first one, which is depicted in \cref{fig:reach_top_border_forward_back_cmd_A}, where we can identify five regions.

In the first region, where $\theta_{c_{p_{1}}}\leq\theta_{l_{L}}$, $\theta_{l_{L_{\text{out}}}}$ is the minimum and $\theta_{l_{R_{\text{out}}}}$ is the maximum. 

\begin{figure}[t]
\centering
\begin{subfigure}[t]{0.16\textwidth}
\centering
\includegraphics[width=1\textwidth]{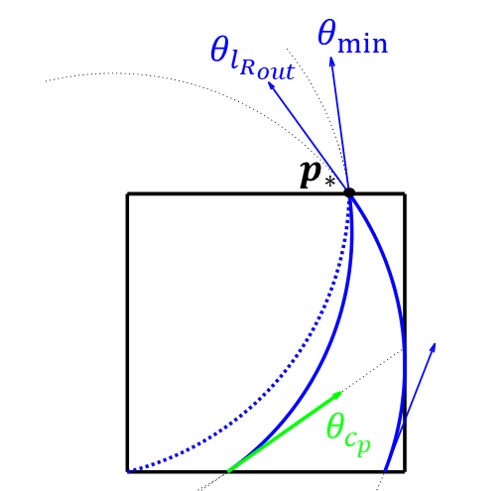}
\caption{ }\label{fig:top_forward_min_limited}
\end{subfigure}\hfill
\begin{subfigure}[t]{0.16\textwidth}
\centering
\includegraphics[width=1\textwidth]{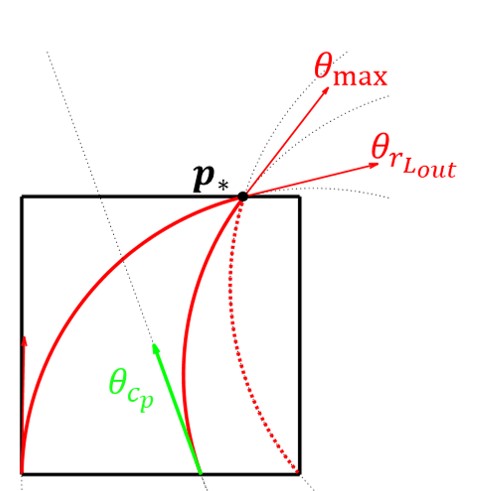}
\caption{ }\label{fig:top_forward_max_limited}
\end{subfigure}\hfill
\begin{subfigure}[t]{0.16\textwidth}
	\centering	\includegraphics[width=1\textwidth]{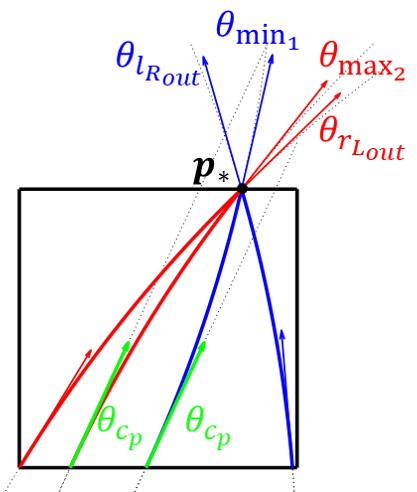}
		\caption{}\label{fig:top_forward_two_regions}
\end{subfigure}
\caption{Example of minimum (a) and maximum (b) limited by $\theta_{c_{p}}$ . Example of forwad set consisting of two continuous regions (c).}\label{fig:top_forward_max_limited_and_two}
\end{figure}

In the second interval, the condition is $\theta_{l_{R}}<\theta_{c_{p_{2}}}\leq\theta_{l_{L}}$. In this case, the maximum is still $\theta_{l_{L_{\text{out}}}}$, while the minimum is computed using the geometrical construction depicted in \cref{fig:top_forward_min_limited}. In particular, we have to find the circle having the following properties:
\begin{enumerate}
\item it intersects the top border in the considered point $p_{*}$;
\item it is tangential to the straight-line having slope $m=\tan(\theta_{c_{p_{2}}})$;
\item it intersects the bottom border in any point.
\end{enumerate}
The minimum corresponding to this circle in our local reference frame is computed as follows:
\begin{equation}
\theta_{\text{min}}=-\asin\left(\frac{r\sin(\theta_{c_{p}}+\frac{\pi}{2})-d}{r}\right)+\frac{\pi}{2}
\label{eq:theta_min_limited_forward}
\end{equation} 

The cases $\theta_{c_{p_{5}}}\geq\theta_{r_{R}}$ and 
$\theta_{r_{L}}<\theta_{c_{p_{4}}}\leq\theta_{r_{R}}$
are dual to the previous two. More specifically, in the first one the minimum is $\theta_{r_{R_{\text{out}}}}$, while the maximum is $\theta_{r_{L_{\text{out}}}}$. Instead, for $\theta_{c_{p_{4}}}$ in order to compute the maximum we must use the circle that intersects the top border in $p_{*}$ and that is tangential to the straight line with slope $m=\tan(\theta_{c_{p_{4}}})$ as shown in \cref{fig:top_forward_max_limited}. In this case, using the local reference frame the formula for the maximum is the following:
\begin{equation}
\theta_{\text{max}}=\asin\left(\frac{r\sin(\theta_{c_{p}}-\frac{\pi}{2})-d}{r}\right)+\frac{\pi}{2}
\label{eq:theta_max_limited_forward}
\end{equation}

The last case that we must consider is $\theta_{l_{L}}<\theta_{c_{p_{3}}}<\theta_{r_{L}}$. In this scenario, the forward reachable set for the considered point is an empty set because neither left nor right turns starting from the bottom border can reach the top border in the considered point. 

Although we mentioned that for backward commands the forward reachable set can consist of two disjoint sets, we can notice that for the case of $\theta_{l_{R}}\leq\theta_{r_{L}}$ we always have a continuous set.
\begin{figure}[t]
	\centering	\includegraphics[width=0.32\textwidth]{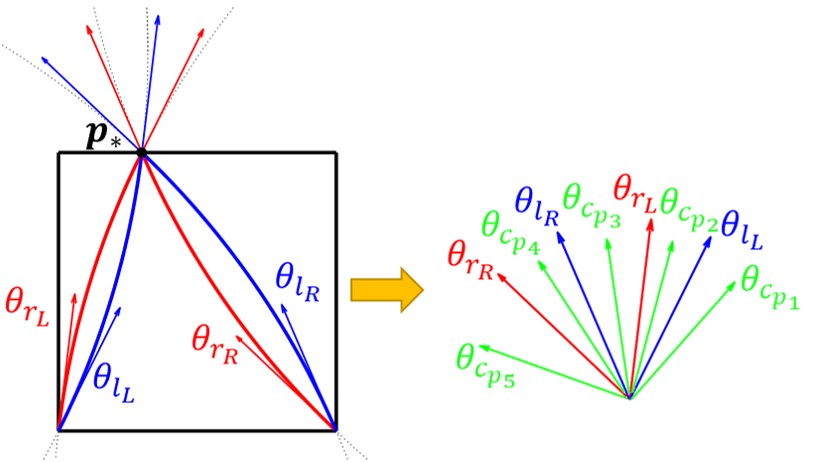}
		\caption{Angle regions for $\theta_{l_{R}}\geq\theta_{r_{L}}$.}\label{fig:reach_top_border_forward_back_cmd_B}
\end{figure}

When $\theta_{l_{R}}>\theta_{r_{L}}$ we can identify again five regions as shown in \cref{fig:reach_top_border_forward_back_cmd_B}. For the cases $\theta_{c_{p_{1}}}$, $\theta_{c_{p_{2}}}$, $\theta_{c_{p_{4}}}$, $\theta_{c_{p_{5}}}$ the forward reachable set is a continuous set delimited by the same minimum and maximum computed for their homonyms in the case of $\theta_{l_{R}}\leq\theta_{r_{L}}$. 

For $\theta_{r_{L}}<\theta_{c_{p_{3}}}<\theta_{l_{R}}$, in general we have two disjoint sets, which are computed using the geometric construction depicted in \cref{fig:top_forward_two_regions}. The first set is delimited by maximum and minimum given by left turns, while in the second set minimum and maximum are given by right turns. In the first case, the maximum is $\theta_{l_{R_{\text{out}}}}$, while the minimum is given by a left turn having circle tangential to the straight line having slope $m=\tan(\theta_{c_{p_{3}}})$. The derivation of the second set is dual. More specifically, in this case the minimum is $\theta_{r_{L_{\text{out}}}}$, while the maximum is obtained with a right turn whose circle is tangential to the straight line having slope $m=\tan(\theta_{c_{p_{3}}})$.

\subsection{Bottom Border Forward Reachable Set at the Right Border: Right Command}
In this section, we consider trajectories that start from the bottom border and end in the right border. We start with the case of right commands, where the boundary of the reachable set is obtained comparing $\theta_{c}$ with the exit angles. Using \cref{eq:circle_2_points} we can find the circles intersecting the edges $e_{L}=(0,0)$ and $e_{R}=(d,0)$ and the considered point $p_{*}=(d,y_{*})$. 

\begin{figure}[t]
    \centering
    \begin{subfigure}{0.16\textwidth}
    \centering
        \includegraphics[width=1\textwidth]{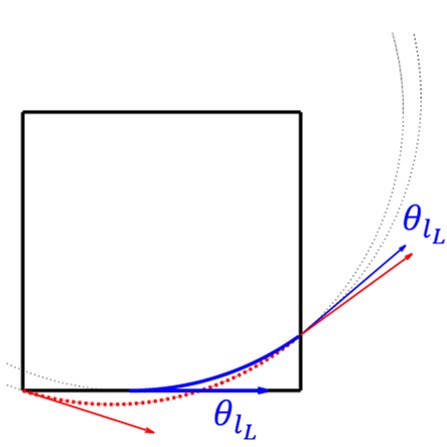}
        \caption{ }
        \label{fig:forward_right_left_turn_tang_bottom}
    \end{subfigure}
	\quad
    \begin{subfigure}{0.16\textwidth}
    \centering
        \includegraphics[width=1\textwidth]{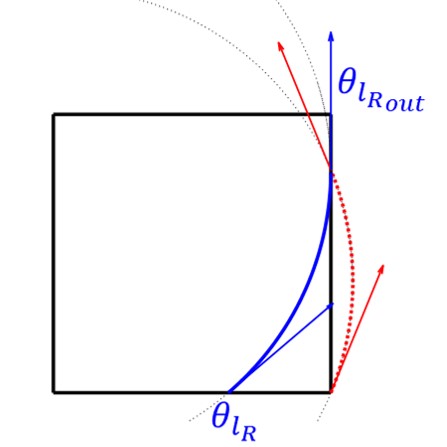}
        \caption{ }
        \label{fig:forward_right_left_turn_tang_right}
    \end{subfigure}
    \caption{Example of left turn tangential to the bottom border (\cref{fig:forward_right_left_turn_tang_bottom}) and the right border (\cref{fig:forward_right_left_turn_tang_right}).}\label{fig:tang_turns_forward_right}
\end{figure}

Also in this case, we must assure that trajectories do not exit from any border before reaching $p_{*}$. In particular, in two cases we must recompute the turns:
\begin{itemize}
\item if in a left turn starting from $e_{L}$ the initial angle is $\theta_{l_{L}}\leq 0$, the circle tangential to the bottom border and intersecting $p_{*}$ must be computed;
\item left turns starting from the right extremity are always tangential to the right border as shown in \cref{fig:forward_right_left_turn_tang_right}, therefore for every point $\theta_{l_{R_{\text{out}}}} = \frac{\pi}{2}$.  
\end{itemize}

When deriving the angle limits, we must distinguish between the two cases: $\theta_{r_{R_{\text{out}}}}<\theta_{l_{L_{\text{out}}}}$ and $\theta_{r_{R_{\text{out}}}}\geq \theta_{l_{L_{\text{out}}}}$.
\begin{figure}[t]
\centering
\begin{subfigure}[t]{0.21\textwidth}
\centering
\includegraphics[width=1\textwidth]{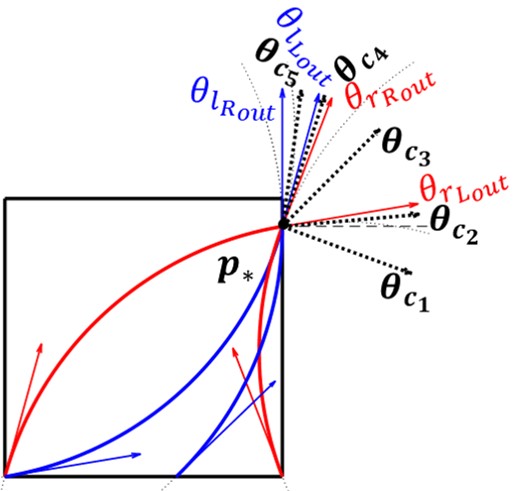}
\caption{ }\label{fig:reach_right_border_right_cmd_A}
\end{subfigure}\hfill
\begin{subfigure}[t]{0.21\textwidth}
\centering
\includegraphics[width=1\textwidth]{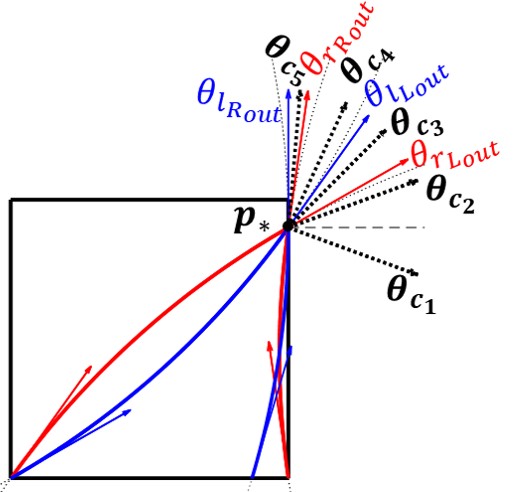}
\caption{ }\label{fig:reach_right_border_right_cmd_B}
\end{subfigure}
\caption{Angle intervals for: $\theta_{r_{R_{\text{out}}}}<\theta_{l_{L_{\text{out}}}}$(a) and $\theta_{r_{R_{\text{out}}}}\geq \theta_{l_{L_{\text{out}}}}$ (b).}\label{fig:reach_right_border_right_cmd}
\end{figure}

\begin{figure}[t]
\centering
\begin{subfigure}[t]{0.18\textwidth}
	\centering
	\includegraphics[width=1\textwidth]{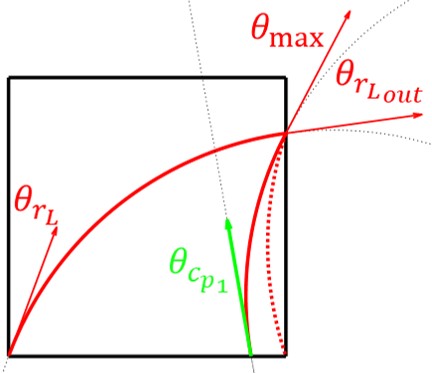}
	\caption{ }			\label{fig:right_forward_max_limited}
\end{subfigure}
\quad
\begin{subfigure}[t]{0.18\textwidth}
\centering	\includegraphics[width=1\textwidth]{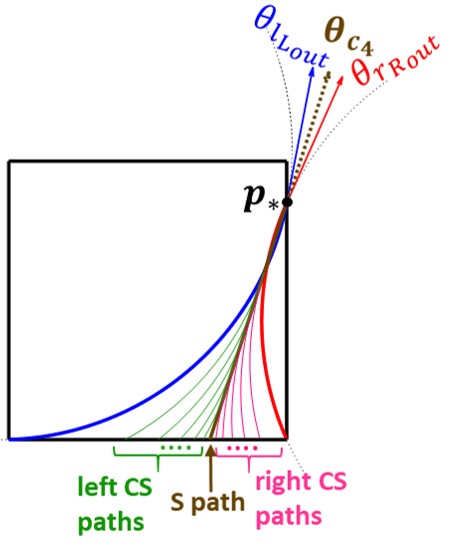}
	\caption{ }\label{fig:forward_right_CS_sets}
\end{subfigure}
\caption{Example of maximum limited by $\theta_{c_{p_{1}}}$ (a). Example of forward reachable set at the right border consisting in a singleton (b).}\label{fig:forward_right_special}
\end{figure}

The case $\theta_{r_{R_{\text{out}}}}<\theta_{l_{L_{\text{out}}}}$ is illustrated in \cref{fig:reach_right_border_right_cmd_A}, where we can identify five intervals for the commanded heading angle. In the first interval the condition is $\theta_{c_{1}}\leq0$. In this case, the minimum is $\theta_{r_{L_{\text{out}}}}$, while for the maximum we must compare $\theta_{r_{R}}$ with $\theta_{c_{p_{1}}}=\theta_{c_{1}}+\pi$. If $\theta_{c_{p_{1}}}\geq\theta_{r_{R}}$, then the minimum is $\theta_{r_{R_{\text{out}}}}$, otherwise it is computed using the circle tangential to the straight line having slope $m=\tan(\theta_{c_{p_{1}}})$ as illustrated in \cref{fig:right_forward_max_limited}.
In this case the maximum is computed as follows:
\begin{equation}
\theta_{\text{max}}=\asin\left(\frac{r\sin(\theta_{c_{p}}-\frac{\pi}{2})-y_{*}}{r}\right)+\frac{\pi}{2}
\label{eq:theta_max_limited_forward_right}
\end{equation}

The second interval is the one in which $0<\theta_{c_{2}}\leq\theta_{r_{L_{\text{out}}}}$. In this case, minimum and maximum are respectively $\theta_{r_{L_{\text{out}}}}$ and $\theta_{r_{R_{\text{out}}}}$. 

In the third interval the commanded heading angle is $\theta_{r_{L_{\text{out}}}}<\theta_{c_{3}}\leq\theta_{r_{R_{\text{out}}}}$, therefore the maximum is still $\theta_{r_{R_{\text{out}}}}$, while the minimum is $\theta_{c_{3}}$ and it is given by RS paths (and an S path if the straight line having slope $m=\tan(\theta_{c_{3}})$ intersects the bottom border). 

In the fourth interval, the forward reachable set in the considered point is a singleton because the condition $\theta_{r_{R_{\text{out}}}}<\theta_{c_{4}}<\theta_{l_{L_{\text{out}}}}$ does not allow the existence of right and left turns that can reach the right border. Instead, as shown in \cref{fig:forward_right_CS_sets}, the considered point $p_{*}$ is reached by an S path and CS paths with final heading equal to $\theta_{c_{4}}$.

The last case is $\theta_{c_{5}}\geq\theta_{l_{L_{\text{out}}}}$, where the minimum is $\theta_{l_{L_{\text{out}}}}$, while the maximum is equal to $\theta_{c_{5}}$ and is given by S and CS paths.

When $\theta_{r_{R_{\text{out}}}}>\theta_{l_{L_{\text{out}}}}$ minimum and maximum for the angles $\theta_{c_{1}}$, $\theta_{c_{2}}$, $\theta_{c_{3}}$, and $\theta_{c_{5}}$ depicted in \cref{fig:reach_right_border_right_cmd_B}, are the ones computed for the homonym angles in the case $\theta_{r_{R_{\text{out}}}}\leq\theta_{l_{L_{\text{out}}}}$. For the fourth interval, since $\theta_{l_{L_{\text{out}}}}<\theta_{c_{4}}<\theta_{r_{R_{\text{out}}}}$, the minimum is $\theta_{l_{L_{\text{out}}}}$ and the maximum is $\theta_{r_{R_{\text{out}}}}$.

\subsection{Bottom Border Forward Reachable Set at the Right Border: Left Command}
We now consider the case $\theta_{c}\in(\frac{\pi}{2},\frac{3}{2}\pi)$. In this scenario, we must compare the starting angles with $\theta_{c_{p}}=\theta_{c}+\pi$ making the distinction between the two cases: $\theta_{r_{L}}\geq\theta_{l_{R}}$ and $\theta_{r_{L}}<\theta_{l_{R}}$. In the first case, the forward reachable set is either a continuous set or an empty set. 
\begin{figure}[t]
	\centering	\includegraphics[width=0.3\textwidth]{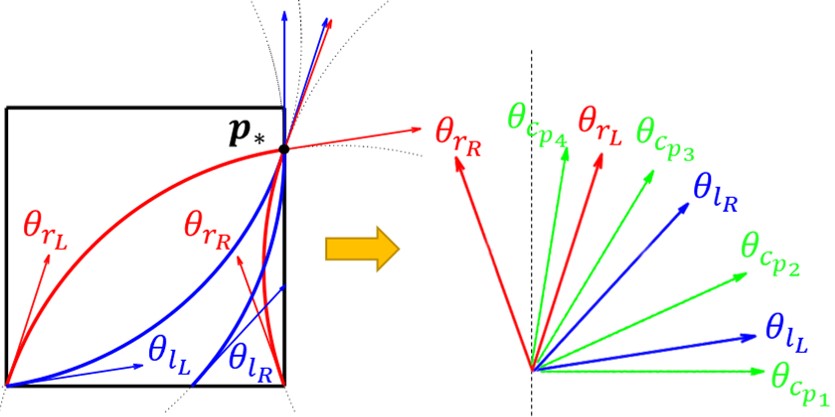}
		\caption{Angle regions for $\theta_{l_{R}}<\theta_{r_{L}}$.}\label{fig:reach_right_border_forward_back_cmd_A}
\end{figure}
In order to compute minimum and maximum we identify four intervals as shown in \cref{fig:reach_right_border_forward_back_cmd_A}. In the first one, since $\theta_{c_{p_{1}}}<\theta_{l_{L}}$, $\theta_{l_{L_{\text{out}}}}$ is the minimum and $\theta_{l_{R_{\text{out}}}}$ is the maximum. 

\begin{figure}[t]
\centering
\begin{subfigure}[t]{0.14\textwidth}
\centering
\includegraphics[width=1\textwidth]{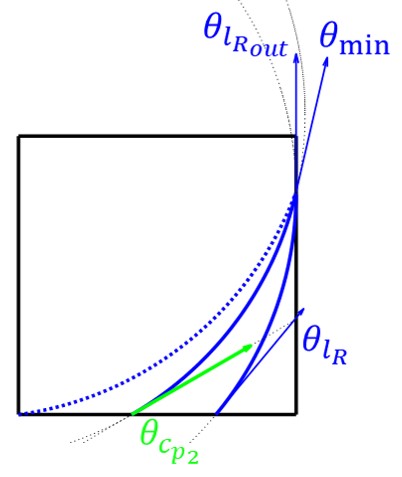}
\caption{ }\label{fig:right_forward_min_limited}
\end{subfigure}\quad
\begin{subfigure}[t]{0.15\textwidth}
\centering
\includegraphics[width=1\textwidth]{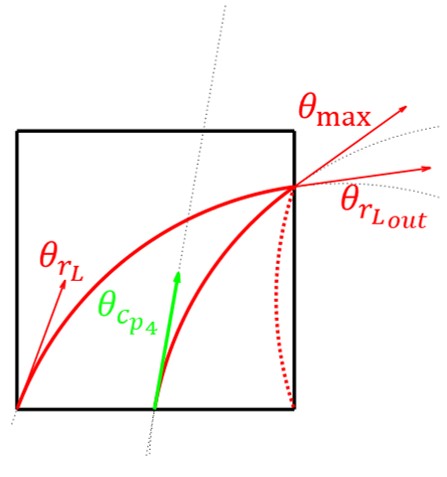}
\caption{ }\label{fig:right_forward_max_limited_2}
\end{subfigure}
\begin{subfigure}[t]{0.17\textwidth}
	\centering	\includegraphics[width=1\textwidth]{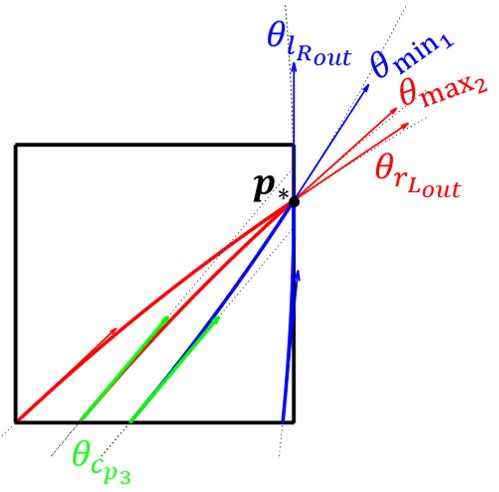}
		\caption{}\label{fig:reach_right_two_regions}
\end{subfigure}
\caption{Example of minimum (a) and maximum (b) limited by $\theta_{c_{p}}$. Example of forward reachable set consisting in two disjoint sets (c).}\label{fig:right_forward_limited}
\end{figure}

When $\theta_{l_{L}}<\theta_{c_{p_{2}}}\leq\theta_{l_{R}}$, $\theta_{l_{R_{\text{out}}}}$ is still the maximum, while the minimum is limited by $\theta_{c_{p_{2}}}$. In particular, we use the geometric construction shown in \cref{fig:right_forward_min_limited} and the minimum is computed as follows:
\begin{equation}
\theta_{\text{min}}=-\asin\left(\frac{r\sin(\theta_{c_{p}}+\frac{\pi}{2})-y_{*}}{r}\right)+\frac{\pi}{2}
\label{eq:theta_min_limited_forward_right}
\end{equation}  

For $\theta_{l_{R}}<\theta_{c_{p_{3}}}<\theta_{r_{L}}$ there is no solution because there not exits a C path that can reach the considered point.

The last case is $\theta_{r_{L}}\geq\theta_{c_{p_{4}}}$, for which $\theta_{r_{L_{\text{out}}}}$ is the minimum, while the maximum is computed using the circle tangential to the straight line having slope $m=\tan(\theta_{c_{p_{4}}})$ as shown in \cref{fig:right_forward_max_limited_2}\footnote{$\theta_{c_{p_{4}}}$ for left commands cannot be greater than $\frac{\pi}{2}$.}. In this case the maximum is computed with \cref{eq:theta_max_limited_forward_right}.
\begin{figure}[t]
	\centering	\includegraphics[width=0.33\textwidth]{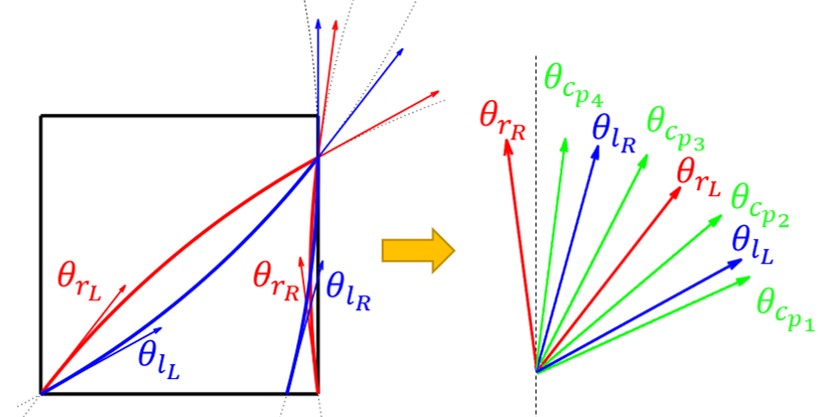}
		\caption{Angle regions for $\theta_{l_{R}}\geq\theta_{r_{L}}$.}\label{fig:reach_right_border_forward_back_cmd_B}
\end{figure}

When $\theta_{r_{L}}<\theta_{l_{R}}$ minimum and maximum for the cases $\theta_{c_{p_{1}}}$, and $\theta_{c_{p_{4}}}$ shown in \cref{fig:reach_right_border_forward_back_cmd_B} are computed as described above for the case $\theta_{r_{L}}\geq\theta_{l_{R}}$. 

\begin{figure}[t]
\centering
\begin{subfigure}[t]{0.2\textwidth}
    \centering
        \includegraphics[width=1\textwidth]{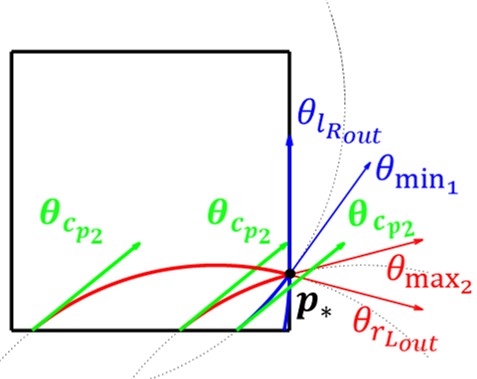}
        \caption{}
        \label{fig:double_region2_right_forward}
\end{subfigure}\\
\begin{subfigure}[t]{0.35\textwidth}
	\centering	\includegraphics[width=1\textwidth]{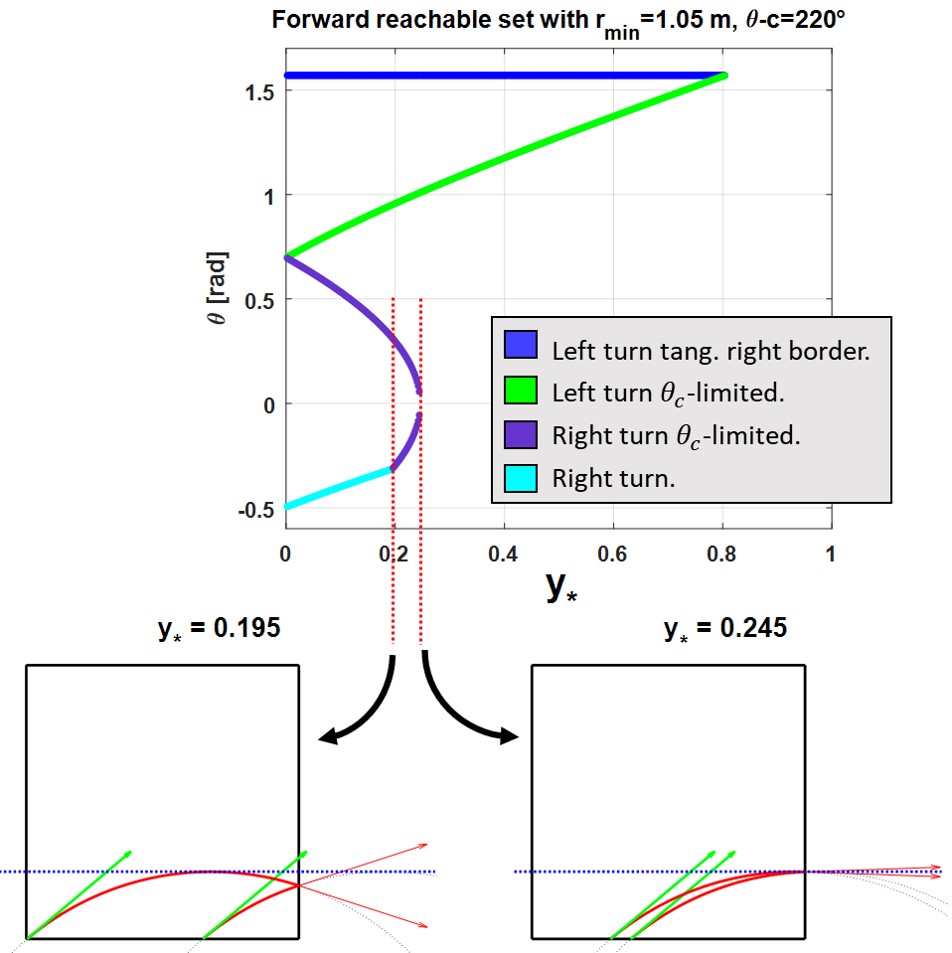}
		\caption{ }\label{fig:example_forwar_right_2_region_2D}
\end{subfigure}
\label{fig:example_2_region_2D_particular}
\caption{Forward reachable set in a single point for $\theta_{l_{L}}<\theta_{c_{p_{2}}}<\theta_{r_{L}}$ (a). Forward reachable set for the whole right border, with focus on the region of right turns starting with $\theta_{l_{L}}(x)<\theta_{c_{p_{2}}}<\theta_{r_{L}}(x)$ (b).}
\end{figure}

For $\theta_{l_{L}}<\theta_{c_{p_{2}}}<\theta_{r_{L}}$, the forward reachable set in $p_{*}$ can be either a continuous set or consist of two disjoint sets. In particular, there is always the set with maximum $\theta_{l_{R_{\text{out}}}}$ and minimum computed with \cref{eq:theta_min_limited_forward_right}. The condition for the existence of the second set is $y_{*}\leq y_{\text{max}}$, with $y_{\text{max}}$ computed as follows:
\begin{equation}
y_{\text{max}}=r\left(1+\sin\left(\theta_{c_{p_{2}}}-\frac{\pi}{2}\right)\right)
\label{eq:y_max}
\end{equation}
In \cref{fig:double_region2_right_forward} there is an example depicting the scenario described above, in which for the point $p_{*}$ there are two disjoint sets and the second one is a set of right turns that start with initial angle $\theta_{c_{p_{2}}}$. 
In \cref{fig:example_forwar_right_2_region_2D} , there is an example of 2-D forward reachable set at the right border, where the horizontal axis is the $y$ position, while the vertical axis is the orientation $\theta$. The upper region of the forward reachable set is given by left turns. Instead in the lower region, the portion on left of the first vertical dashed line corresponds to the case of $\theta_{c_{p_{3}}}$ of \cref{fig:reach_right_border_forward_back_cmd_B}, while the region between the two lines corresponds to $\theta_{c_{p_{2}}}$ of \cref{fig:reach_right_border_forward_back_cmd_B}. For this portion of the lower region, the figure illustrates also the trajectories for the extreme points.

Also in the case $\theta_{r_{L}}<\theta_{c_{p_{3}}}<\theta_{l_{R}}$ we can have two disjoint sets as illustrated in \cref{fig:reach_right_two_regions}. In this case, in the first set, the maximum is $\theta_{l_{R_{\text{out}}}}$, while the minimum $\theta_{\text{min}_{1}}$ is computed with \cref{eq:theta_min_limited_forward_right}. In the second set, the minimum is $\theta_{r_{L_{\text{out}}}}$, while the maximum $\theta_{\text{max}_{2}}$ is computed with \cref{eq:theta_max_limited_forward_right}. 

\section{GLOBAL BACKWARD REACHABILITY}
In the previous sections, we first illustrated how to compute the backward reachable set of a border $l_{i}$ in a generic point. Then we mentioned that fixing one of the coordinates, we obtain the function specific for points lying on a border (different from $l_{i}$). Subsequently, we illustrated how to compute the forward reachable set of a border $l_{i}$ in points lying on a border $l_{j}$ with $i\neq j$. In this section, we illustrate how to combine the two types of sets to compute the global backward reachable set of a region of interest $W_{*}$. 

The strength of this method is that rather than performing a backpropagation of the region in the 3-D space, we expand the borders. The advantages are twofold: reduced computational complexity and reduced spatial complexity. The computational complexity is reduced because rather than discretizing the whole 3-D space and evolving the trajectories, we evolve the boundaries of the cells. The memory usage is also reduced because the reachability of a region is encoded with a set of 2-D maps rather than a 3-D map.

\subsection{Backward Border Mapping}
The pivotal mechanism for the computation of the global backward reachable set is the propagation of a region using the intersection of the regions delimited by $\Theta(x,\theta_{c})$ and $\Phi(x,\theta_{c})$.
\begin{figure}[t]
\begin{subfigure}[t]{0.2\textwidth}
	\centering
	\includegraphics[width=1\textwidth]{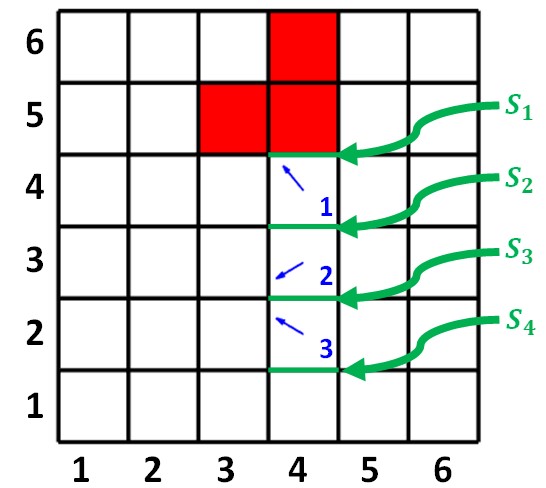}
		\caption{ }\label{fig:example_work_space}
\end{subfigure}
\begin{subfigure}[t]{0.26\textwidth}
	\centering
	\includegraphics[width=1\textwidth]{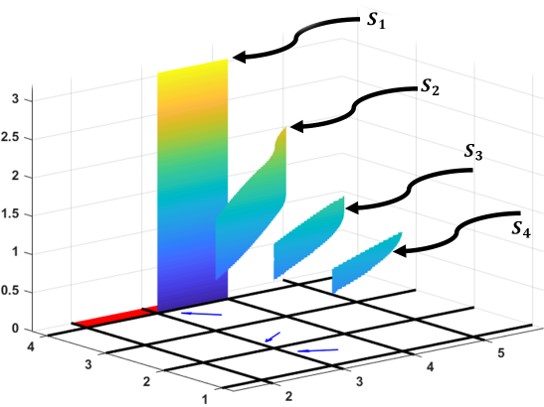}
		\caption{ }\label{fig:example_work_space_expansion}
\end{subfigure}
\caption{Example of workspace with region of interst (red cells) (a). 3-D view of backward expansion in one direction.}\label{fig:example_work_application}
\end{figure}
\begin{figure}[t]
\centering
\begin{subfigure}[t]{0.23\textwidth}
	\centering
	\includegraphics[width=1\textwidth]{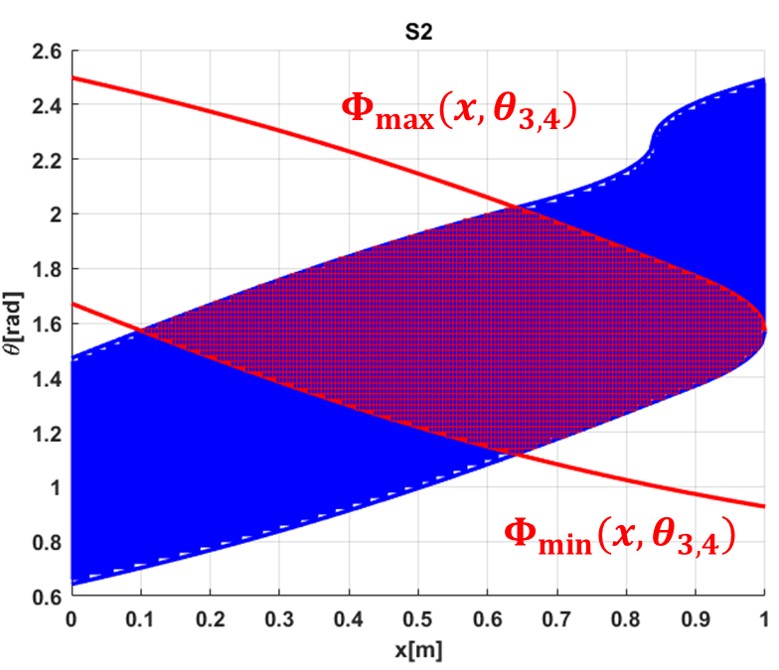}
		\caption{ }\label{fig:example_intersection}
\end{subfigure}
\begin{subfigure}[t]{0.23\textwidth}
	\centering
	\includegraphics[width=1\textwidth]{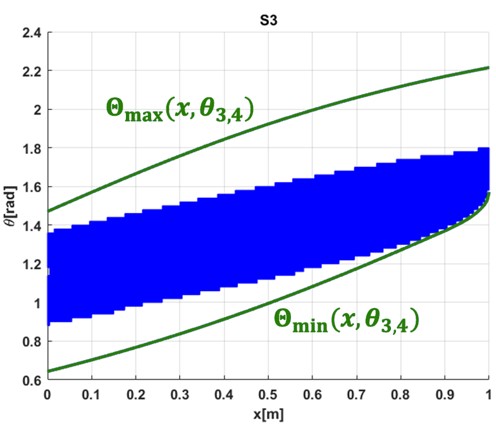}
		\caption{ }\label{fig:example_subset}
\end{subfigure}
		\caption{Intersection between region of interest backward reachable set (blue area) and forward reachable set (red curves) (a). Set $S_{3}$ is a subset of the cellular backward reachable set (delimited by green curves) (b).}\label{fig:example_propagation_step}
\end{figure}
To illustrate this process we consider the example shown in \cref{fig:example_work_application}, where
for the sake of clarity, we illustrate the backward expansion only for one side and only in one direction, therefore the figures shown in the following explanation show ``incomplete'' sets. Referring to \cref{fig:example_work_space}, we consider the backward expansion of the bottom border of the cell (5,4)\footnote{In the notation ($i$,$j$), $i$ is the row and $j$ is the column.} downwards across the lower three cells neglecting the expansion through the side borders.

We start with the set $S_{1}$ which is in one of the borders of the region of interest, therefore $S_{1}\equiv[0,d]\times[0,2\pi]$ because any configuration is considered part of the backward reachable set. In order to propagate the set $S_{1}$ downwards to the bottom border of cell (4,4), we must use $\Theta_{B_{\text{min}}}^{T}(x,\theta_{c_{4,4}})$ and $\Theta_{B_{\text{max}}}^{T}(x,\theta_{c_{4,4}})$, where $B$ and $T$ means that trajectories start from the bottom border and end in the top border. These two functions provide the limits of the set $S_{2}$, which contains all the initial configurations of the bottom border of the cell (4,4) from which the top border of the same cell is reachable.

The next step is the propagation of the set $S_{2}$ towards the bottom border of cell (3,4). The set delimited by $\Theta_{B_{\text{min}}}^{T}(x,\theta_{c_{3,4}})$ and $\Theta_{B_{\text{max}}}^{T}(x,\theta_{c_{3,4}})$ contains all the starting configurations from which the top border of cell (3,4) is reachable. However, among those configurations, we are only interested in those for which trajectories end in $S_{2}$. Therefore, $S{3}$ consists of all the configurations whose integration leads to the intersection between the forward reachable set delimited by $\Phi_{B_{\text{min}}}^{T}(x,\theta_{c_{3,4}})$ and $\Phi_{B_{\text{max}}}^{T}(x,\theta_{c_{3,4}})$ and the set $S_{2}$ (\cref{fig:example_intersection}). As illustrated in \cref{fig:example_subset}, $S_{3}$ is a subset of the cellular backward reachable set. In this example, the last step is the propagation of $S_{3}$ to the bottom of cell (2,4), obtaining $S_{4}$. \cref{fig:example_work_space_expansion} shows the complete expansion from $S_{1}$ to $S_{4}$.

\subsection{Iterative Border Expansion}
The propagation of the borders must be computed in every direction. This leads to the presence of many disjoint sets for each border, which we must keep track of during the expansion. As an alternative to the propagation of single borders through the whole map, we can use \cref{alg:complete_map_gen}.
\IncMargin{0.3em}
\begin{algorithm}[t]
\SetKwData{bmap}{GM}
\SetKwData{brs}{BRS}\SetKwData{buffer}{Buffer}
\SetKwData{cost}{Cost}\SetKwData{goals}{$\text{S}_{\text{GC}}$}
\SetKwProg{foreach}{foreach}{ do}{end}
\SetKw{return}{return}
\SetKwInOut{Input}{input}\SetKwInOut{Output}{output}
\Input{$W_{*}$, grid-map \bmap }
\Output{Backward Reachable Set BRS}
\BlankLine
\While{\text{at least one configuration is added to BRS}}{
\foreach{\text{border i}}{
\foreach{\text{neighbor j }}{
\uIf{$\text{neighbor j} \not\in W_{*}$ and $\text{neighbor j} \not\in$ \emph{\bmap margins} }{
propagate sets of i to j
}
}
}}
\return \brs
\caption{Iterative Border Expansion}\label{alg:complete_map_gen}
\end{algorithm}\DecMargin{0.3em}

\begin{figure}[t]
\begin{subfigure}[t]{0.36\textwidth}
	\centering
	\includegraphics[width=1\textwidth]{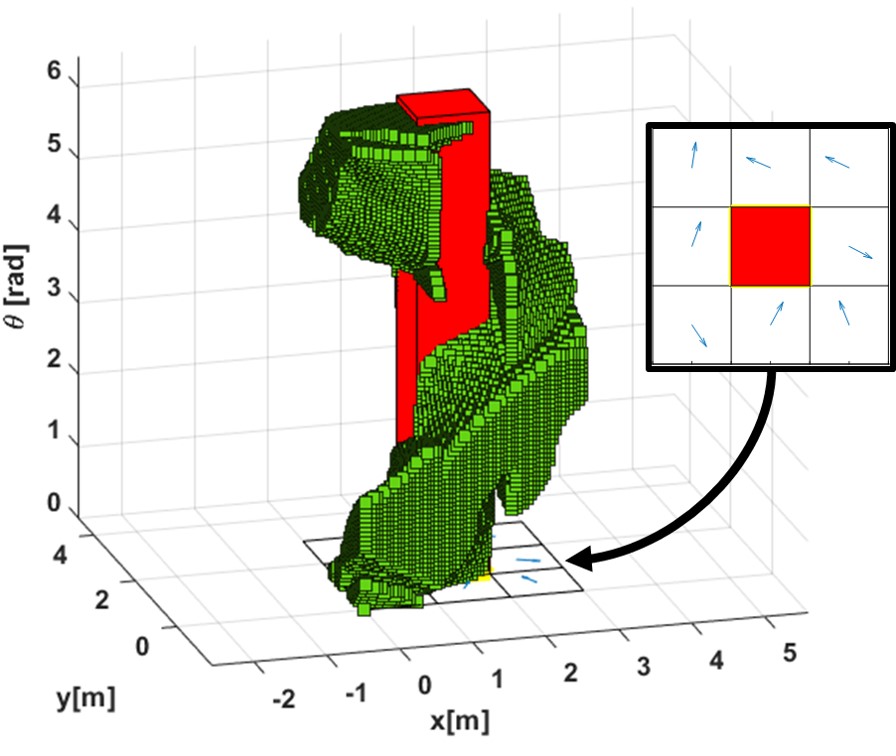}
		\caption{ }\label{fig:complete_back_set}
\end{subfigure}
\begin{subfigure}[t]{0.12\textwidth}
	\centering
	\includegraphics[width=1\textwidth]{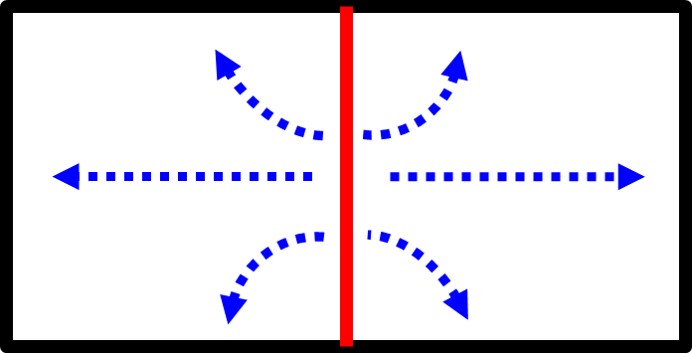}
		\caption{ }\label{fig:six_neigh}
\end{subfigure}
\caption{Global backward reachable set (a). Six neighboring borders(b).}\label{fig:algo_illustration}
\end{figure}

 In this case, for each border, there is a bitmap that indicates which configurations are part of the backward reachable set of the region of interest. In the first iteration, only borders of the region of interest have nonempty sets. For each iteration, the set of a border is propagated to the $6$ neighboring borders as shown in \cref{fig:six_neigh}. The algorithm iterates until no new configuration is marked as part of the backward reachable set and it returns a map for each border of the grid-map. 

The 3-D reachability map can be computed in real-time by checking if the trajectory starting from the considered configuration (internal to the cell) ends in a cell of the border map marked as part of the backward reachable set. An example of complete backward reachable set is shown in \cref{fig:complete_back_set}.  

The space complexity of the algorithm can be evaluated by assuming that the workspace is discretized with a square grid-map having $n$ cells per side. Neglecting the borders at the margins of the grid-map, the number of tables necessary to characterize the whole workspace is at most $2n(n-1)$. Assuming that angle and position on the border are discretized using $m$ cells, the number of bits necessary to represent the backward reachability is $m^{2}(2n(n-1))$. Instead, the discretization of the whole 3-D global map requires $m^{3}n^{2}$ bits. For instance with $n=20$ and $m=200$, the number of required bits in our method is $3.04\cdot10^{7}$, while for a 3-D discretization it is $3.2\cdot10^{9}$.
\section{Conclusions}
In this paper, we presented a verification method that is based on the computation of the backward reachable set of a region of interest. The method is formulated specifically for vehicles having a minimum turning radius that is greater than the resolution of the discrete feedback motion plan that they must follow. 

The paper illustrates the geometric principles used for the computation of the reachable sets and practical implementation of the method based on an iterative expansion of the borders. Future work will focus on the extension of the method to the case of bounded angular acceleration and the development of a more efficient numerical implementation.  
 
\bibliography{bib_file}
\bibliographystyle{IEEEtran}

\end{document}